\documentclass{article}

\usepackage[final]{AFME_neurips2024/neurips_2024}

\usepackage[utf8]{inputenc} 
\usepackage[T1]{fontenc}    
\usepackage[backref=page]{hyperref}       
\usepackage{url}            
\usepackage{booktabs}       
\usepackage{amsfonts}       
\usepackage{nicefrac}       
\usepackage{microtype}      
\usepackage{xcolor}         

\usepackage{caption}
\usepackage{subcaption}
\usepackage{wrapfig}
\usepackage{multirow}

\usepackage{amsmath}
\usepackage{amssymb}
\usepackage{mathtools}
\usepackage{amsthm}

\theoremstyle{plain}
\newtheorem{theorem}{Theorem}[section]
\newtheorem{proposition}[theorem]{Proposition}

\theoremstyle{definition}

\theoremstyle{remark}

\usepackage{bm}
\usepackage{amsfonts}
\newcommand{\vect}[1]{{\bm{#1}}}
\newcommand{\norm}[1]{\left\lVert#1\right\rVert}

\def\frr{\mathrm{FRR}}
\def\trr{\mathrm{TRR}}
\def\far{\mathrm{FAR}}
\def\roc{\mathrm{ROC}}

\def\bfar{{\rm BFAR}}
\def\bfrr{{\rm BFRR}}

\title{Mitigating Bias in Facial Recognition Systems: \\
Centroid Fairness Loss Optimization}

\author{%
  Jean-Rémy Conti \\
  LTCI, Télécom Paris\\
  Institut Polytechnique de Paris\\
  \texttt{jean-remy.conti@telecom-paris.fr}\thanks{Alternative correspondence: \texttt{jeanremy.conti@gmail.com}.} \\
  \And
  Stéphan Clémençon \\
  LTCI, Télécom Paris \\
  Institut Polytechnique de Paris \\
  \texttt{stephan.clemencon@telecom-paris.fr} \\
}

\begin{document}

\maketitle

\begin{abstract}
  The urging societal demand for fair AI systems has put pressure on the research community to develop predictive models that are not only globally accurate but also meet new fairness criteria, reflecting the lack of disparate mistreatment with respect to sensitive attributes (\textit{e.g.} gender, ethnicity, age). In particular, the variability of the errors made by certain Facial Recognition (FR) systems across specific segments of the population compromises the deployment of the latter, and was judged unacceptable by regulatory authorities. Designing fair FR systems is a very challenging problem, mainly due to the complex and functional nature of the performance measure used in this domain (\textit{i.e.} ROC curves) and because of the huge heterogeneity of the face image datasets usually available for training. In this paper, we propose a novel post-processing approach to improve the fairness of pre-trained FR models by optimizing a regression loss which acts on centroid-based scores. Beyond the computational advantages of the method, we present numerical experiments providing strong empirical evidence of the gain in fairness and of the ability to preserve global accuracy.
\end{abstract}

{\bf Publishing note.} 
The present work has been accepted at the "Algorithmic Fairness through the Lens of Metrics and Evaluation" (AFME) Workshop at NeurIPS 2024. A preliminary version (14 pages) of the present paper was accepted at the "Fairness in Biometrics" Workshop at ICPR 2024, resulting in a publication within the book "Pattern Recognition. ICPR 2024 International Workshops and Challenges", Volume~15614 of the Lecture Notes in Computer Science series, by Springer Nature.

\section{Introduction}
\label{sec:introduction}

Facial Recognition (FR) systems are increasingly deployed (\textit{e.g.} at border checkpoints), for biometric verification in particular. Although the global accuracy attained by certain FR systems 
is now judged satisfactory and offers clear efficiency gains \citep[see e.g.][]{krizhevsky2012imagenet}, their operational deployment has revealed statistically significant disparities in treatment between different segments of the population. Fairness in algorithmic decisions is now a major concern and is becoming part of the functional specifications of AI systems, and soon subject to regulation in certain application domains, including FR \citep{frvt_part8}.
Following recent scandals\footnote{See e.g. this \href{https://www.aclu.org/blog/privacy-technology/surveillance-technologies/amazons-face-recognition-falsely-matched-28}{study} conducted by the American Civil Liberties Union that attracted notable media attention.}, the academic community has delved into the investigation of bias in FR systems in the last few years. This exploration extends back to early studies which examined racial bias in non-deep FR models \citep{nist_report_2002}. A recent comprehensive analysis conducted by the U.S. National Institute of Standards and Technology (NIST) unveiled significant performance disparities among hundreds of academic/commercial FR algorithms, based on \textit{e.g.} gender \citep{frvt_part8}. In the present work, fairness is understood as the absence of (significant) disparate mistreatment, and we propose a novel methodology to reduce such bias in deep learning-based FR systems.



\looseness=-1 \textbf{Related work on mitigating bias in FR.} Various approaches have been explored to address bias in deep learning: pre-processing, in-processing, and post-processing methods \citep{caton2020fairness}. These strategies differ based on whether the fairness intervention occurs before, during, or after the training phase. 
Pre-processing methods
are deemed unsuitable for FR purposes since balanced training datasets are actually {\it not enough} to mitigate bias, as illustrated by \cite{gender_balanced_data} for gender bias and \cite{race_balanced_data} for racial bias. In terms of in-processing, 
\citet{mitigating_bias_RL} use reinforcement learning for fair decision rules, but face computational challenges. \citet{alasadi2019toward} and \citet{debface} employ adversarial methods to reduce bias, but these are recognized for their instability and computational needs, while \citet{rfw} leverage imbalanced and transfer learning techniques. Note that in-processing strategies require a complete retraining, which is notoriously costly as state-of-the-art FR systems require very large training datasets. Furthermore, these strategies lead to fairness improvements at the expense of the performance, highlighting a performance-fairness trade-off \citep{du2020fairness}. Concerning post-processing approaches,
\citet{pass} mitigate the racial bias of a pre-trained model by enforcing the embeddings not to contain any racial information. \citet{conti_vmf} reduce the gender bias using a statistical model for the embedding space, but they admit that the method is not able to tackle other types of bias. Those works change the pre-trained embeddings to improve the fairness, both in terms of false positives and false negatives. In contrast, another line of research takes a different approach, not altering the latent space but modifying the decision rule itself. \citet{terhorst2020post} intervene on the score function, while \citet{faircal} rely on calibration methods. Those works focus on the bias in terms of false positives and their training set needs to have the same distribution than the test set, which may not be a realistic scenario.

\looseness=-1 \textbf{Contributions.} We propose a post-processing approach to mitigate the bias of a pre-trained/frozen FR model that is accessible only as a black-box, making it applicable to numerous already deployed FR systems. It is important to note that many methods fine tune state-of-the-art open-source FR models, thereby acquiring their bias, which underscores the necessity of improving their fairness properties. Our solution aims to align intra-group performance curves with those of a reference group, a functional objective that is inherently challenging.
By drawing an analogy between real FR scores and centroid-based scores, we simplify the original fairness objective, enabling the use of pseudo-metrics that are easier to compute and align with modern FR loss functions. This approach thus bridges the gap between contemporary FR training and fairness mitigation. We introduce a new loss function, Centroid Fairness, which aligns these pseudo-metrics across the subgroups of the population, and use this loss to train a small model called the Fairness Module. Extensive experiments demonstrate that our Fairness Module reduces bias in pre-trained models while maintaining their performance, surpassing the traditional performance-fairness trade-off. In summary, our bias mitigation method eliminates the need for a costly retraining of a large FR model, is especially fast to train, and retains the state-of-the-art performance of the pre-trained model.




\section{Background and Preliminaries}
\label{sec:background}

FR mainly serves two use cases: \textit{identification}, involving the recognition of the identity of a probe image among several pre-enrolled identities, and \textit{verification} (the primary focus of this paper), aiming at deciding whether two face images correspond to the same identity. Facial verification operates in an open-set scenario: the identities present at the test phase are often absent from the training set.

\textbf{Notations.} The indicator function of an event $\mathcal{E}$ is denoted by $\mathbb{I}\{\mathcal{E}\}$. Assuming that there are $K\leq +\infty$ identities within the images, a FR dataset of size $N$ is denoted by $(\vect{x}_i, y_i)_{1 \leq i \leq N}$, where $\vect{x}_i \in \mathbb{R}^{h \times w \times c}$ is a face image of size $h \times w$, $c$ is the color channel dimension and $y_i \in \{1, \ldots, K \}$ is the identity label of $\vect{x}_i$.
In face verification, the standard approach is to train an encoder function $f_\theta: \mathbb{R}^{h \times w \times c} \rightarrow \mathbb{R}^d$ (\textit{e.g.} CNN) with learnable parameters~$\theta$ to bring images from the same identity closer together and images from distinct identities far away from each other in $\mathbb{R}^d$. The latent representation of an image $\vect{x}_i$ is referred to as its \textit{face embedding} $f_\theta(\vect{x}_i)$.



\subsection{Evaluation of Face Recognition Systems}\label{subsec:performance_FR}


We start by explaining how the performance of a trained FR model $f_\theta$  
is evaluated.

\textbf{Decision rule.}  
The \textit{similarity score} between two face images $\vect{x}_i, \vect{x}_j$ is usually measured using the cosine similarity between their embeddings:
\begin{align}\label{eq:cosine_sim}
    s_\theta(\vect{x}_i, \vect{x}_j) &= \cos(f_\theta(\vect{x}_i), f_\theta(\vect{x}_j)) \in [-1,1], 
\end{align}
where $\cos(\vect{z}, \vect{z}') = \vect{z}^\intercal \vect{z}'/(\norm{\vect{z}} \cdot \norm{\vect{z}'})$ and $|| \cdot ||$ is the Euclidean norm.
The decision rule to decide whether both images share the same identity is obtained by applying a threshold $t \in [-1,1]$ to this score. If $s_\theta(\vect{x}_i, \vect{x}_j) > t$, $(\vect{x}_i, \vect{x}_j)$ is predicted to share the same identity (positive pair), while for $s_\theta(\vect{x}_i, \vect{x}_j) \leq t$ we predict that they do not (negative pair). In this sense, face verification is a binary classification with a pair of images as input. 

\textbf{Evaluation metrics.}
The gold standard to evaluate the performance of FR models is the $\roc$ curve, \textit{i.e.} the False Rejection Rate ($\frr$) 
as a function of the False Acceptance Rate ($\far$) 
as the threshold $t$ varies\footnote{Standard $\roc$ definitions often used $1-\frr$ (i.e., the True Positive Rate) instead of $\frr$. The FR community favors the use of $\frr$ so that both metrics $\far$ and $\frr$ correspond to error rates.}.
In practice, $\far$ and $\frr$ are computed on an evaluation dataset $(\vect{x_i}, y_i)_{1 \leq i \leq N}$.
The set of $\textit{genuine}$ (ground-truth positive) pairs is denoted as $\mathcal{G}~=~\{ (\vect{x}_i, \vect{x}_j), 1~\leq~i~<~j~\leq~N,~y_i~=~y_j \}$ while the set of \textit{impostor} (ground-truth negative) pairs is $\mathcal{I}~=~\{ (\vect{x}_i, \vect{x}_j), 1 \leq i < j \leq N, y_i \neq y_j \}$. The metrics $\far$ and $\frr$ of a FR model $f_\theta$ are then given by: 
\begin{align}
 \mathrm{FAR}(t) &= \frac{1}{|\mathcal{I}|} \ \sum_{(\vect{x}_i, \vect{x}_j) \in \mathcal{I}} \mathbb{I}\{ s_\theta(\vect{x}_i, \vect{x}_j) > t \}, \ \ \ \ \ \ 
 \mathrm{FRR}(t) = \frac{1}{|\mathcal{G}|} \ \sum_{(\vect{x}_i, \vect{x}_j) \in \mathcal{G}} \mathbb{I}\{ s_\theta(\vect{x}_i, \vect{x}_j) \leq t \} 
 \label{eq:far&frr},
\end{align}
where $|\cdot|$ denotes the cardinality of a set. $\far(t)$ is the proportion of \textit{impostor scores} 
predicted as positive (same identity), while $\frr(t)$ is the proportion of \textit{genuine scores} 
predicted as negative (distinct identities). The trade-off between these two metrics play a pivotal role in assessing FR systems.
For instance, in the context of airport boarding gates, achieving a very low $\far$ is crucial, while simultaneously maintaining a reasonable $\frr$ to ensure a smooth and user-friendly experience.

The $\roc$ curve, evaluated at $\alpha \in (0,1)$, is naturally defined as $\frr(t)$ with $t$ such that ${\far(t)=\alpha}$. More rigorously, it is defined using the generalized inverse of a distribution function (see Appendix~\ref{app:inverse_far}), as
    $\roc(\alpha) = \frr \big( \ \far^{-1}(\alpha) \ \big)$.
The $\mathrm{FAR}$ level $\alpha$ establishes the operational threshold of the Face Recognition system, representing the acceptable security risk. Depending on the use case, it is typically set to $10^{-i}$ with $i \in \{1, \ldots, 6\}$.

\subsection{Fairness Metrics}\label{subsec:fairness_metrics_FR}
 
\looseness=-1 To assess the fairness of a FR model, the typical approach involves examining differentials in performance across several subgroups or segments of the population. These subgroups are defined by a \textit{sensitive attribute}, such as gender, ethnicity or age class.
For a given discrete sensitive attribute that can take $A > 1$ different values in say $\mathcal{A} = \{0, 1, \ldots, A-1 \}$, the attribute of identity $k \in \{1, \ldots, K\}$ is denoted by $a_k \in \mathcal{A}$.
With those notations, the attribute of a face image $\vect{x}_i$ with identity $y_i$ is thus $a_{ y_i }$.

\textbf{Intra-group metrics.} For any $a\in \mathcal{A}$, the sets of genuine/impostor pairs with attribute $a$ are: 
\begin{align}
    \mathcal{G}_a&=\{ (\vect{x}_i, \vect{x}_j),  i \! < \! j, y_i = y_j, a_{ y_i } \! = \! a_{ y_j } \! = \! a \}, 
    \ \ \ 
    \mathcal{I}_a=\{ (\vect{x}_i, \vect{x}_j),  i \! < \! j , y_i \neq y_j, a_{ y_i } \! = \! a_{ y_j } \! = \! a \}. 
    \label{eq:real_gen&imp_a_pairs}
\end{align}
From $\mathcal{G}_a$ and $\mathcal{I}_a$,
one naturally defines the intra-group metrics $\far_a(t)$ and $\frr_a(t)$ as for Eq.~\eqref{eq:far&frr}, by replacing $\mathcal{I}$ by $\mathcal{I}_a$ and $\mathcal{G}$ by $\mathcal{G}_a$.
Because FR systems use a unique threshold $t$ for their decision rule, our focus will be on comparing intra-group metrics $(\far_a(t))_{a \in \mathcal{A}}$ (and $(\frr_a(t))_{a \in \mathcal{A}}$) at any fixed threshold $t$, as recommended by \citet{robinson2020face} and \citet{krishnapriya2020issues}.

\textbf{Fairness metrics.}
We rely on fairness metrics used in previous work \citep{conti_uncertainty} that align with those used by the NIST in their FRVT report, analyzing the fairness of hundreds of academic/commercial FR models~\citep{frvt_part8}. As the FR performance may be more focused on the $\far$ metric or on $\frr$, depending on the use-case, we consider one fairness measure to quantify the differentials in $(\far_a(t))_{a \in \mathcal{A}}$, and another for $(\frr_a(t))_{a \in \mathcal{A}}$:
\begin{align}    
\textstyle \bfar(t) =  \max_{a \in \mathcal{A}} \far_a(t)/\far^\dag(t), \quad
\bfrr(t) =  \max_{a \in \mathcal{A}} \frr_a(t) / \frr^\dag(t), \label{eq:bfar_bfrr}
\end{align}
where $\far^\dag(t)$ (resp. $\frr^\dag(t)$) is the geometric mean of the values $(\far_a(t))_{a \in \mathcal{A}}$ (resp. $(\frr_a(t))_{a \in \mathcal{A}}$). One can read the above acronyms as ``Bias in $\mathrm{FAR}$/$\mathrm{FRR}$''. Both metrics compare the worst $\far$/$\frr$ performance across subgroups to an aggregated performance over all subgroups, at a fixed threshold $t$. As for the $\roc$ curve, this threshold $t$ is set as $t = \far^{-1}(\alpha)$, with the $\far$ level $\alpha \in (0,1)$ defining the selected appropriate security risk. The $\far$ level $\alpha$ is computed using the global population of the evaluation dataset, and not for some specific subgroup. In this sense, both fairness metrics are in fact functions of the $\far$ level $\alpha$, instead of $t$, as for the $\roc$ curve.

We note that other FR fairness metrics exist in the literature, such as the maximum difference in the values $(\mathrm{FAR}_a(t))_{a \in \mathcal{A}}$ used by \citet{alasadi2019toward} and \citet{pass}. However, this metric is not normalized and thus lacks interpretability.

\subsection{Training a Face Recognition System with Pseudo-Scores}\label{subsec:FR_training}

We now review the state-of-the-art approach to train the deep encoder $f_\theta: \mathbb{R}^{h \times w \times c} \rightarrow \mathbb{R}^d$ on a large 
training set $(\vect{x}_i, y_i)_{1 \leq i \leq N}$ with $K$ identities.
During the training phase exclusively, a fully-connected layer is added on top of the embeddings, resulting in an output of a $K$-dimensional vector that predicts the identity of each image within the training set. The complete model (the encoder and the fully-connected layer) is trained as an identity classification task.
The predominant form of FR loss functions is \citep{normface,vmf_deep_learning}:
\begin{align}
&\mathcal{L}(\theta, \vect{\mu}) = - \frac{1}{N} \sum\limits_{i=1}^N \log \bigg( \frac{ e^{\displaystyle \kappa \ \overline{s}_\theta(\vect{x}_i, \vect{\mu}_{y_i}) }}{\sum_{k=1}^K e^{\displaystyle \kappa \ \overline{s}_\theta(\vect{x}_i, \vect{\mu}_k) }} \bigg), 
 \ \ \ \ \ \text{with }\overline{s}_\theta(\vect{x}_i, \vect{\mu}_k) = \cos(f_\theta(\vect{x}_i), \vect{\mu}_k) 
 ,
\label{eq:pseudo_score&standard_loss}
\end{align} 
where the $\vect{\mu}_k$'s are the fully-connected layer's parameters ($\vect{\mu}_k \in \mathbb{R}^d)$, $\kappa > 0$ is the inverse temperature of the softmax and $\vect{\mu} = ( \vect{\mu}_k )_{1 \leq k \leq K}$. Recent loss functions slightly change $\overline{s}_\theta(\vect{x}_i, \vect{\mu}_{y_i})$ in Eq.~\eqref{eq:pseudo_score&standard_loss} by incorporating a fixed margin to penalize more the intra-class angle variations. Examples include CosFace \citep{cosface} and ArcFace \citep{arcface} and we apply our debiasing method on both models. Note that minimizing the loss in Eq.~\eqref{eq:pseudo_score&standard_loss} enforces the embeddings $f_\theta(\vect{x}_i)$ 
of identity $l$ to be all clustered around $\vect{\mu}_l$ on the unit hypersphere, while being far from any $\vect{\mu}_k$ such that $k \neq l$. In this sense, one might consider $\vect{\mu}_k$ as a \textit{pseudo}-embedding which represents the identity $k$. The $\vect{\mu}_k$'s are often called \emph{centroids} \citep{zhu2019new} 
of a class/identity.

\textbf{Pseudo-scores.} As a consequence of the embedding-like nature of the centroids, we call $\overline{s}_\theta(\vect{x}_i, \vect{\mu}_k)$ in Eq.~\eqref{eq:pseudo_score&standard_loss} the \textit{pseudo-score} between image $\vect{x}_i$ and centroid $\vect{\mu}_k$. Denoting the image-centroid genuine/impostor pairs by
\begin{align}
    \overline{\mathcal{G}}&=\{ (\vect{x}_i, \vect{\mu}_k), \
    y_i = k \},
    \ \ \ \ \ \ \ \ \  
    \overline{\mathcal{I}}=\{ (\vect{x}_i, \vect{\mu}_k), \
    y_i \neq k \}, 
    \label{eq:pseudo_gen&imp_pairs}
\end{align}
\looseness=-1 we distinguish between \textit{genuine pseudo-scores} $\{ \overline{s}_\theta(\vect{x}_i, \vect{\mu}_k), (\vect{x}_i, \vect{\mu}_k) \in \overline{\mathcal{G}}\}$ and \textit{impostor pseudo-scores} $\{ \overline{s}_\theta(\vect{x}_i, \vect{\mu}_k), (\vect{x}_i, \vect{\mu}_k) \in \overline{\mathcal{I}}\}$. 
These pseudo-scores effectively serve as a proxy for real scores during training. As  the number of pseudo-scores is much smaller than the number of real scores, this makes the centroid-based approach much more efficient than prior strategies relying on real scores, like triplet loss-based approaches \citep{facenet}. Our Fairness Module defined in the next section builds upon this strategy to preserve efficiency in the post-processing step.

\section{Centroid Fairness}
\label{sec:centroid_fairness}

\looseness=-1 We now present our approach. We assume to have access to a pre-trained FR model $f$, trained on a training set $(\vect{x}_i, y_i)_{1 \leq i \leq N}$. We consider here a \emph{black-box setting}, where $f$ is only available for inference and its training parameters are unavailable (hence we drop $\theta$ from the notation $f_\theta$). Face images are encoded by $f$ into embeddings, and scores (Eq.~\ref{eq:cosine_sim}) are computed for all embedding pairs of the test set. Those scores define $\far_a(t)$, $\frr_a(t)$ metrics for the pre-trained model $f$, for each subgroup $a \in \mathcal{A}$, and the $\far_a(t)$ curves are typically not aligned/equal when $a$ varies (and similarly for $\frr_a(t)$). The objective is to improve the fairness metrics of $f$ (as defined in Eq.~\ref{eq:bfar_bfrr}). Our post-processing method consists in training a small fair model acting on the output of $f$.
In practice, we use the same training set as $f$ for fair comparisons (see Appendix~\ref{app:reason_reuse_train_set} for a discussion).


\textbf{Pseudo-scores estimation.} Our approach uses the pseudo-scores of $f$, but assuming access to the pre-trained centroids $\vect{\mu}_k$ 
is not realistic as many open-source pre-trained FR models do not provide those parameters. 
Following Section~\ref{subsec:FR_training}, 
a natural way to estimate the centroid of identity $k$ for $f$ consists in taking the average of all normalized embeddings $f(\vect{x}_i)$ of identity $k$:
\begin{equation}\label{eq:PT_centroids}
   \textstyle \vect{\mu}_k^{(0)} = \frac{1}{n_k} \sum_{i=1}^N  \mathbb{I}\{y_i =k\}\frac{f(\vect{x}_i)}{\norm{f(\vect{x}_i)}} \in \mathbb{R}^d,
\end{equation}
where $n_k$ is the number of images $\vect{x}_i$ from the training set whose identity satisfies $y_i = k$. The (estimated)  \textit{pre-trained} pseudo-scores $\overline{s}(\vect{x}_i, \vect{\mu}_k^{(0)})$ can then be computed 
using Eq.~\eqref{eq:pseudo_score&standard_loss}.

\subsection{Pseudo-Score Transformation }\label{subsec:alignement_pseudo_metrics}

As seen in Section~\ref{subsec:fairness_metrics_FR}, to achieve fairness for a trained model $f$, the intra-group metrics $(\far_a(t))_{a \in \mathcal{A}}$ and $(\frr_a(t))_{a \in \mathcal{A}}$ should be nearly constant when $a \in \mathcal{A}$ varies, at any threshold $t \in [-1,1]$. Those intra-group metrics are computed with real scores $s(\vect{x}_i, \vect{x}_j)$. 

\textbf{Pseudo-metrics.} Motivated by the analogy between real scores $s(\vect{x}_i, \vect{x}_j)$ and pseudo-scores $\overline{s}(\vect{x}_i, \vect{\mu}_k^{(0)})$ in Section~\ref{subsec:FR_training}, we define centroid versions of the intra-group metrics. The real intra-group metrics involve image pairs $(\vect{x}_i, \vect{x}_j)$ which belong to $\mathcal{G}_a$ or to $\mathcal{I}_a$ (see Eq.~\ref{eq:real_gen&imp_a_pairs}), \textit{i.e.} genuine/impostor pairs sharing the same attribute $a \in \mathcal{A}$. Extending the definitions of the image-centroid pairs $\overline{\mathcal{G}}$ and $\overline{\mathcal{I}}$ in Eq.~\eqref{eq:pseudo_gen&imp_pairs}, we define their counterparts for image-centroid pairs sharing the same attribute $a \in \mathcal{A}$:
\begin{align}
    \overline{\mathcal{G}}_a =\{ (\vect{x}_i, \vect{\mu}_k^{(0)}), y_i = k, a_{ y_i } = a_{ k } = a \}, \quad
    \overline{\mathcal{I}}_a =\{ (\vect{x}_i, \vect{\mu}_k^{(0)}), y_i \neq k, a_{ y_i } = a_{ k } = a \}.
\end{align}
One can then naturally define the intra-group \textit{pseudo-metrics} for the pre-trained model $f$:
\begin{align}
 \overline{\far}_a(t) = 
 1/|\overline{\mathcal{I}}_a|
 \! \! \! \! \! \! \! \! \! \sum_{(\vect{x}_i, \vect{\mu}_k^{(0)}) \in \overline{\mathcal{I}}_a} 
 \! \! \! \! \! \! \! 
 \mathbb{I}\{ \overline{s}(\vect{x}_i, \vect{\mu}_k^{(0)}) > t \}, \ \ \ \
 \overline{\frr}_a(t) = 
 1/|\overline{\mathcal{G}}_a|
  \! \! \! \! \! \! \! \! \!
 \sum_{(\vect{x}_i, \vect{\mu}_k^{(0)}) \in \overline{\mathcal{G}}_a}
 \! \! \! \! \! \! \! 
 \mathbb{I}\{ \overline{s}(\vect{x}_i, \vect{\mu}_k^{(0)}) \leq t \} 
 \label{eq:pseudo_far&frr_a}.
\end{align}
Just like pseudo-scores $\overline{s}(\vect{x}_i, \vect{\mu}_k^{(0)})$ are surrogates for real scores $s(\vect{x}_i, \vect{x}_j)$, intra-group pseudo-metrics $\overline{\far}_a$ and $\overline{\frr}_a$ are surrogates for intra-group real metrics $\far_a$ and $\frr_a$. Those pseudo-metrics are displayed in Fig. \ref{fig:ps_frr} and \ref{fig:ps_far} for the FR model ArcFace (solid lines), evaluated on the BUPT dataset which is annotated with race labels $a$ (see Section~\ref{sec:experiments} for more details). For the same model and the same dataset, we show the real metrics $\far_a$ and $\frr_a$ in Fig. \ref{fig:rs_frr} and \ref{fig:rs_far} (solid lines). We see that pseudo and real metrics behave similarly, and the ranking of the best performance among groups is preserved. We will thus work with intra-group pseudo metrics for efficiency purposes.

\textbf{Pseudo-metric curve alignment.} 
Our objective is to align the curves $(\overline{\far}_a(t))_{a \in \mathcal{A}}$ across the values of $a$, for any $t$, and similarly for $(\overline{\frr}_a(t))_{a \in \mathcal{A}}$. 
For simplicity, let us consider two subgroups: $a$ and $r$, where  $r$ is the \textit{reference} subgroup on which one would like to align all other subgroups. In this case, the objective is to align the curve $\overline{\far}_a(t)$ on $\overline{\far}_r(t)$, and $\overline{\frr}_a(t)$ on $\overline{\frr}_r(t)$. 

\begin{wrapfigure}{r}{0.5\textwidth}
\begin{center}
\vskip -.6cm
\centerline{\includegraphics[width=0.5\columnwidth]{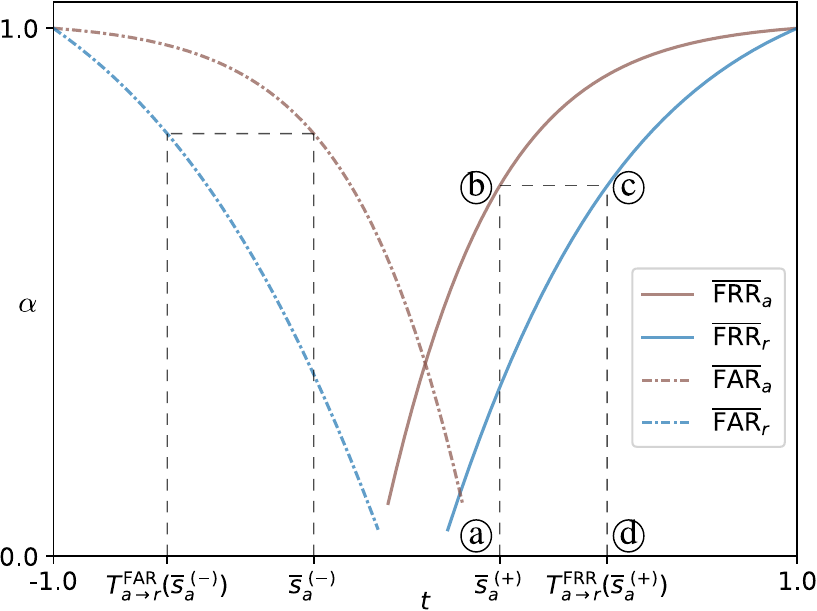}}
\caption{Pseudo-score transformation to achieve fairness. From a pseudo-score $\overline{s}_a^{(-)}$ (resp. $\overline{s}_a^{(+)}$) of an image-centroid impostor (resp. genuine) pair sharing the attribute $a$, the pseudo-metric $\overline{\far}_a(\overline{s}_a^{(-)}) = \alpha$ (resp. $\overline{\frr}_a(\overline{s}_a^{(+)})=\alpha$) is computed. The transformed score is the score which makes the reference pseudo-metric $\overline{\far}_r$ (resp. $\overline{\frr}_r$) equal to $\alpha$, among the scores from image-centroid pairs of attribute $r$.}
\label{fig:schema_score_transfo}
\vskip -2cm
\end{center}
\end{wrapfigure}

\looseness=-1 To explain how the transformation works, we rely on Figure~\ref{fig:schema_score_transfo}. Let $(\vect{x}_i, \vect{\mu}_k^{(0)}) \in \overline{\mathcal{G}}_a$ be an image-centroid genuine pair with attribute $a$, to which the pre-trained model $f$ assigns the pseudo-score ${\overline{s}_a^{(+)} := \overline{s}(\vect{x}_i, \vect{\mu}_k^{(0)})}$ shown in \textcircled{a}. $\overline{s}_a^{(+)}$ is only involved in the computation of $\overline{\frr}_a(t)$ of the pre-trained model $f$ (see Eq.~\ref{eq:pseudo_far&frr_a}). Let $\alpha := \overline{\frr}_{a} [\overline{s}_a^{(+)}]$ be the $\overline{\frr}_{a}$ metric evaluated at $\overline{s}_a^{(+)}$, shown in \textcircled{b}. By definition of $\overline{\frr}_a$, $\overline{s}_a^{(+)}$ is the $\alpha \text{-th}$ quantile of the genuine pseudo-scores of attribute $a$ (\textit{i.e.}~of~${\{ \overline{s}(\vect{x}_i, \vect{\mu}_k^{(0)}): (\vect{x}_i, \vect{\mu}_k^{(0)}) \in \overline{\mathcal{G}}_a\}}$). To have $\overline{\frr}_a(t)$ aligned with $\overline{\frr}_r(t)$, it suffices that their respective quantiles are equal. The $\alpha \text{-th}$ quantile of the genuine pseudo-scores of attribute $r$ (\textit{i.e.}~of~${\{ \overline{s}(\vect{x}_i, \vect{\mu}_k^{(0)}) : (\vect{x}_i, \vect{\mu}_k^{(0)}) \in \overline{\mathcal{G}}_r\}}$), is attained for a certain pseudo-score $\overline{s}_{\text{target}}$ satisfying $\overline{\frr}_{r} [\overline{s}_{\text{target}}] = \alpha$ (shown in \textcircled{c}). Using the quantile function $(\overline{\frr}_{r})^{-1}$, it means that ${\overline{s}_{\text{target}} = (\overline{\frr}_{r})^{-1}(\alpha)}$, as shown in~\textcircled{d}. 

Thus, for all $(\vect{x}_i, \vect{\mu}_k^{(0)}) \in \overline{\mathcal{G}}_a$, we define the pseudo-score transformation for $\overline{s}_a^{(+)} :=  \overline{s}(\vect{x}_i, \vect{\mu}_k^{(0)})$ as
\begin{equation*}
T_{a \xrightarrow[]{} r}^{\frr}(\overline{s}_a^{(+)}) = (\overline{\frr}_{r})^{-1} \circ \overline{\frr}_{a} [\overline{s}_a^{(+)}]. 
\end{equation*}
Similarly, for $(\vect{x}_i, \vect{\mu}_k^{(0)}) \in \overline{\mathcal{I}}_a$, we define the pseudo-score transformation for $\overline{s}_a^{(-)} :=  \overline{s}(\vect{x}_i, \vect{\mu}_k^{(0)})$ as
\begin{equation*}
   T_{a \xrightarrow[]{} r}^{\far}(\overline{s}_a^{(-)}) = (\overline{\far}_{r})^{-1} \circ \overline{\far}_{a} [\overline{s}_a^{(-)}]. 
\end{equation*}

The notation $\circ$ denotes the composition operator, and we refer to Appendix~\ref{app:inverse_pseudo_far_frr} for rigorous definitions of the quantiles $\overline{\far}_{r}^{-1}$ and $\overline{\frr}_{r}^{-1}$.
The transformations $T_{a \xrightarrow[]{} r}^{\far}$, $T_{a \xrightarrow[]{} r}^{\frr}$ are illustrated in Fig.~\ref{fig:schema_score_transfo}. Assume that all the pseudo-scores of group $a$ are modified as follows: all pairs from $\overline{\mathcal{I}}_a$ are transformed with $T_{a \xrightarrow[]{} r}^{\far}$ while all pairs from $\overline{\mathcal{G}}_a$ are transformed with $T_{a \xrightarrow[]{} r}^{\frr}$. This operation changes the pseudo-metrics $\overline{\far}_a(t)$ and $\overline{\frr}_a(t)$. As detailed in Appendix~\ref{app:effect_ps_transformation}, and proved with Proposition~\ref{prop:theory_alignment}, the newly obtained pseudo-metrics are aligned with $\overline{\far}_r$ and $\overline{\frr}_r$ respectively. In other words, $T_{a \xrightarrow[]{} r}^{\far}$ is an impostor pseudo-score transformation which aligns $\overline{\far}_a(t)$ on $\overline{\far}_r(t)$, and $T_{a \xrightarrow[]{} r}^{\frr}$ is a genuine pseudo-score transformation which aligns $\overline{\frr}_a(t)$ on $\overline{\frr}_r(t)$.

\subsection{Fairness Module}

Leveraging the pseudo-score transformations which align the group-wise pseudo-metrics with the pseudo-metrics of a reference group presented in Section~\ref{subsec:alignement_pseudo_metrics}, we are now ready to present the Fairness Module which will be responsible for learning these transformations.

\begin{figure}[t]
\vskip 0.2in
\begin{center}
\centerline{\includegraphics[width=0.8\columnwidth]{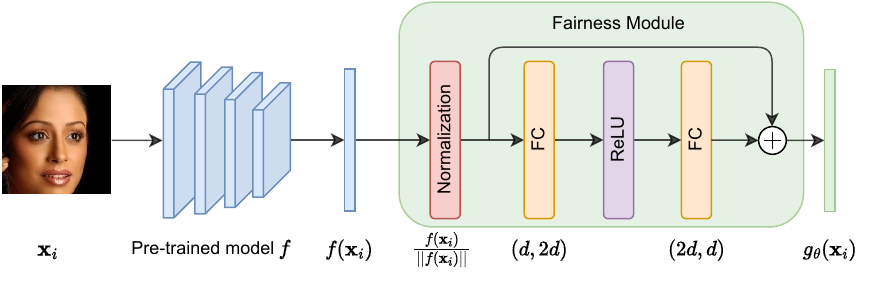}}
\caption{The proposed Fairness Module framework. A frozen pre-trained model~$f$ outputs the embedding $f(\vect{x}_i)$ for the image $\vect{x}_i$. The Fairness Module outputs a new fair embedding $g_\theta(\vect{x}_i)$.}
\label{fig:framework}
\end{center}
\vskip -0.3in
\end{figure}

\looseness=-1 The Fairness Module $g_\theta$ with parameters $\theta$ takes as input an embedding $f(\vect{x}_i) \in \mathbb{R}^d$ of the pre-trained model $f$ and outputs a new embedding $g_\theta(\vect{x}_i)\in \mathbb{R}^d$ of the same dimension (see Fig.~\ref{fig:framework}). To prioritize simplicity and scalability, the architecture is a shallow MLP of size ($d$, $2d$, $d$), with the first layer followed by a ReLU activation.
Motivated by the normalization of FR embeddings, both for FR training and FR evaluation (see Section~\ref{sec:background}), we add a normalization step before entering the MLP. We also add a shortcut connection after the normalization step in order to allow for the model to fit the identity function easily: by setting all the MLP's weights to $0$, the Fairness Module outputs the same embeddings than $f$. As gains in FR fairness are often accompanied by high losses in performance \citep{pass},
this shortcut connection makes it easy for the module to recover the performance of $f$ if needed.


To learn the pseudo-score transformation introduced in Section~\ref{subsec:alignement_pseudo_metrics}, the Fairness Module needs to output new pseudo-scores. 
Thus, in addition to the new embeddings $g_\theta(\vect{x}_i)$, we learn $K$ new centroids $\vect{\mu}_k$, as detailed in the next section. Since the shortcut connection makes it easy for the Fairness Module to recover the embedding space of the pre-trained model $f$, we initialize those new centroids $\vect{\mu}_k$ with the pre-trained centroids $\vect{\mu}_k^{(0)}$. Thus, the parameters used to train the Fairness Module are: the MLP's weights $\theta$ and the $K$ new centroids $\vect{\mu}_k$.  
The new pseudo-scores obtained from the Fairness Module are then given by
$\overline{s}_\theta(\vect{x}_i, \vect{\mu}_k) = \cos(g_\theta(\vect{x}_i), \vect{\mu}_k) \in [-1,1]$.
Note that we use the notation $\overline{s}_\theta(\vect{x}_i, \vect{\mu}_k)$ for the pseudo-scores of the Fairness Module, while the pseudo-scores of the pre-trained model $f$ are denoted by $\overline{s}(\vect{x}_i, \vect{\mu}_k^{(0)}) = \cos(f(\vect{x}_i), \vect{\mu}_k^{(0)})$.

\subsection{Centroid Fairness Loss}\label{subsec:CF_loss}

We now present the loss function used to train the Fairness Module, following the pseudo-score transformation of Section~\ref{subsec:alignement_pseudo_metrics}. Recall that we have access to estimated pre-trained centroids $\vect{\mu}_k^{(0)}$, the pre-trained pseudo-scores $\overline{s}(\vect{x}_i, \vect{\mu}_k^{(0)})$ and thus the (pre-trained) pseudo-metrics  $\overline{\far}_a$, $\overline{\frr}_a$, for any $a \in \mathcal{A}$. A reference group $r \in \mathcal{A}$ is chosen, on which to align the pseudo-metrics of all groups.

\looseness=-1 The loss function we propose formulates the problem as a regression task over the Fairness Module pseudo-scores $\overline{s}_\theta(\vect{x}_i, \vect{\mu}_k)$, where the target for this regression are given by transformed pre-trained pseudo-scores.
More precisely, let us consider an impostor image-centroid pair $(\vect{x}_i, \vect{\mu}_k^{(0)})\in \overline{\mathcal{I}}_{a_k}$ sharing the same attribute $a_{y_i} = a_k$. The pre-trained model assigned this pair the pseudo-score $\overline{s}(\vect{x}_i, \vect{\mu}_k^{(0)})$. As seen in Section~\ref{subsec:alignement_pseudo_metrics}, to achieve fairness, the pre-trained model $f$ should have given the pseudo-score 
\begin{equation*}
T_{a_k \xrightarrow[]{} r}^{\far}(\overline{s}(\vect{x}_i, \vect{\mu}_k^{(0)})) = (\overline{\far}_{r})^{-1} \circ \overline{\far}_{a_k} [ \overline{s}(\vect{x}_i, \vect{\mu}_k^{(0)})]    
\end{equation*}
 to this pair. Therefore, we use this as the target score for the corresponding pseudo-score $\overline{s}_\theta(\vect{x}_i, \vect{\mu}_k)$ of the Fairness Module.
 In other words, 
 we learn new impostor pseudo-scores such that the $\far$ pseudo-metric of the Fairness Module aligns with the pre-trained pseudo-metric $\overline{\far}_r$ of the reference group~$r$. We use the squared error as loss function for this pair $(\vect{x}_i, \vect{\mu}_k)$: 
\begin{equation*}
    l_{\far}^{(i,k)}(\theta, \vect{\mu}) = \big[ \ \overline{s}_\theta(\vect{x}_i, \vect{\mu}_k) - T_{a_k \xrightarrow[]{} r}^{\far}(\overline{s}(\vect{x}_i, \vect{\mu}_k^{(0)})) \ \big]^2.
\end{equation*}
The same idea follows for a genuine image-centroid pair $(\vect{x}_i, \vect{\mu}_k^{(0)}) \in \overline{\mathcal{G}}_{a_k}$. The pre-trained model $f$ gave the pseudo-score $\overline{s}(\vect{x}_i, \vect{\mu}_k^{(0)})$ and thus the target score is:
\begin{equation*}
T_{a_k \xrightarrow[]{} r}^{\frr}(\overline{s}(\vect{x}_i, \vect{\mu}_k^{(0)})) = (\overline{\frr}_{r})^{-1} \circ \overline{\frr}_{a_k} [ \overline{s}(\vect{x}_i, \vect{\mu}_k^{(0)})], 
\end{equation*}
and the loss function for this pair is then:
\begin{equation*}
    l_{\frr}^{(i,k)}(\theta, \vect{\mu}) = \big[ \ \overline{s}_\theta(\vect{x}_i, \vect{\mu}_k) - T_{a_k \xrightarrow[]{} r}^{\frr}(\overline{s}(\vect{x}_i, \vect{\mu}_k^{(0)})) \ \big]^2.
\end{equation*}
Note that the target scores of our regression task are fully determined by the pre-trained model $f$, and thus fixed before the training of the Fairness Module.
Finally, we combine these pair-level loss functions into two \emph{weighted} global losses, one for the $\far$ bias and another for the $\frr$ bias:
\begin{align*}
    \mathcal{L}_{\far}(\theta,\vect{\mu}) = & \bigg( \sum_{\substack{1 \leq j \leq N \\ 1 \leq l \leq K}}  w_\far^{(j,l)} \bigg)^{-1} \sum_{\substack{1 \leq i \leq N \\ 1 \leq k \leq K}}  w_\far^{(i,k)} \ l_{\far}^{(i,k)}(\theta, \vect{\mu}),\\
    \mathcal{L}_{\frr}(\theta, \vect{\mu}) = &\bigg( \sum_{\substack{1 \leq j \leq N \\ 1 \leq l \leq K}}  w_\frr^{(j,l)} \bigg)^{-1} \sum_{\substack{1 \leq i \leq N \\ 1 \leq k \leq K}} w_\frr^{(i,k)} \ l_{\frr}^{(i,k)}(\theta, \vect{\mu}).
\end{align*}
We define the weights $w_\far^{(i,k)}$ and $w_\frr^{(i,k)}$ given to $l_{\far}^{(i,k)}$ and $l_{\frr}^{(i,k)}$ as:
\begin{align}
    w_\far^{(i,k)} = \frac{\mathbb{I}\{y_i \neq k\} \mathbb{I}\{a_{y_i} = a_k\}}{ |\overline{\mathcal{I}}_{a_k}| \ \overline{\far}_{a_k}[\overline{s}(\vect{x}_i, \vect{\mu}_k^{(0)})]}, \quad
    w_\frr^{(i,k)} = \frac{\mathbb{I}\{y_i = k\} \mathbb{I}\{a_{y_i} = a_k\}}{ |\overline{\mathcal{G}}_{a_k}| \ \overline{\frr}_{a_k}[\overline{s}(\vect{x}_i, \vect{\mu}_k^{(0)})]}.
\end{align}
These weights are carefully chosen to match our objective. First, 
as seen above, $l_{\far}^{(i,k)}$ is only used for pairs in $\overline{\mathcal{I}}_{a_k}$, and $l_{\frr}^{(i,k)}$ for pairs in $\overline{\mathcal{G}}_{a_k}$. Thus, it is necessary to enforce ${w_\far^{(i,k)} \propto \mathbb{I}\{y_i \neq k\} \mathbb{I}\{a_{y_i} = a_k\}}$ and $w_\frr^{(i,k)} \propto \mathbb{I}\{y_i = k\} \ \mathbb{I}\{a_{y_i} = a_k\}$.
Second, the diversity of FR use-cases requires to have good performance and fairness at all $\far$ and $\frr$ levels, especially at very low levels. We thus seek to 
give the same importance to all $\far$ and $\frr$ levels. To account for the fact that the number of pairs in a given FRR or FAR interval is proportional to the length of this interval, we weight each pair $(\vect{x}_i, \vect{\mu}_k^{(0)})$ by a factor $\overline{\far}_{a_k}[\overline{s}(\vect{x}_i, \vect{\mu}_k^{(0)})])^{-1}$ (if it is an impostor) or $\overline{\frr}_{a_k}[\overline{s}(\vect{x}_i, \vect{\mu}_k^{(0)})])^{-1}$ (if it is genuine). Without this weighting, $90\%$ of the non-zero terms of the loss $\mathcal{L}_\frr$ would be devoted to achieve fairness at thresholds $t$ associated to $\frr$ levels in $(10^{-1}, 10^0]$, by definition of $\overline{\frr}_{a}(t)$.  Third, we seek to give the same weight to all groups, irrespective of the number of images for each group. As groups are typically imbalanced in real datasets, we divide by a factor $|\overline{\mathcal{I}}_{a_k}|$ the weight of impostor pairs with attribute $a_k$, and by a factor $|\overline{\mathcal{G}}_{a_k}|$ the weight of genuine pairs with attribute $a_k$.
A more detailed discussion on the choice of those weights can be found in Appendix~\ref{app:weights_discussion}.


Finally, we define the Centroid Fairness objective as the sum of the two previous loss functions, each responsible for either $\far$ fairness or $\frr$ fairness: 
\begin{equation}\label{eq:CF_loss}
    \mathcal{L}_{\mathrm{CF}}(\theta, \vect{\mu}) = \mathcal{L}_{\far}(\theta, \vect{\mu}) + \mathcal{L}_{\frr}(\theta, \vect{\mu}).
\end{equation}
\textbf{Limitations.} Our method is specific to post-processing as it needs some pre-trained information and it may be promising to define a new reference/target arbitrarily for in-processing. In addition, the $\mathcal{L}_{\mathrm{CF}}$ loss requires the attribute labels for the training set only. This is a small limitation as there exists such public datasets and the FR fairness literature builds on them. Moreover, current attribute predictors \citep{fairface_predictor}, from face inputs, achieve $\sim 95\%$ accuracy on popular benchmarks. In terms of computation, the method is efficient as the Fairness Module is shallow and does not depend on the complexity of the pre-trained model (see \ref{app:eval_other_pt_models}).
The method takes $\sim 2.5$ less training time when dividing the train set size by $2$ (see \ref{app:eval_buptef}).

\section{Numerical Experiments}
\label{sec:experiments}

\textbf{Datasets.} The training set considered in this paper is BUPT-Globalface \citep{bupt}. It contains $2$M face images from $38$k celebrities and is annotated with race attributes: ${\mathcal{A}=\{\text{African, Asian, Caucasian, Indian}\}}$. According to \citet{bupt}, its racial distribution is approximately the same as the real distribution of the world’s population, thus an imbalanced and realistic training set. Using the attributes in $\mathcal{A}$, we tackle the racial FR bias. Although other sensitive attributes could be relevant, we choose to mitigate the racial bias, as BUPT is the largest public FR training set, labelled with sensitive attributes (race labels only).
The evaluation of all models is achieved with the test set RFW \citep{rfw}. RFW contains $40$k images from $11$k identities and is widely used to evaluate the racial bias of FR models as it is balanced in images and identities, across the four races in $\mathcal{A}$. To have good estimates of the fairness metrics $\bfar$ and $\bfrr$ at low $\far$ levels $\alpha$, all image pairs, sharing a same race, are used. We also employ the FairFace dataset \citep{fairface} as test set. It was introduced at the ECCV 2020 FairFace challenge and we use their predefined image pairs for evaluation. Contrary to RFW, FairFace is annotated with binary skintone attributes (Dark, Bright) and we inspect the skintone bias with $\bfar$ and $\bfrr$. All face images are resized to $(h,w,c)=(112,112,3)$ pixels with RetinaFace \citep{retinaface}.

\begin{figure*}[t]
\centering
\begin{minipage}{.46\textwidth}
  \centering
  \includegraphics[width=0.95\columnwidth]{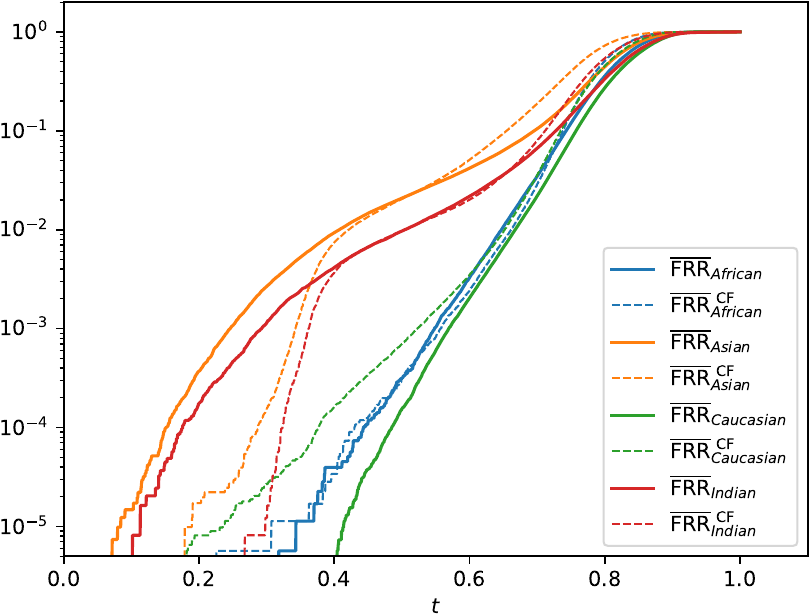}
  \captionof{figure}{Pseudo-metrics $\overline{\frr}_a$ obtained with pseudo-scores, for each race. The pseudo-scores come from the pre-trained model $f$ (solid lines) or the Fairness Module (dashed).}
\label{fig:ps_frr}
\end{minipage}%
\hspace{0.5cm}
\begin{minipage}{.46\textwidth}
  \centering
  \includegraphics[width=0.95\columnwidth]{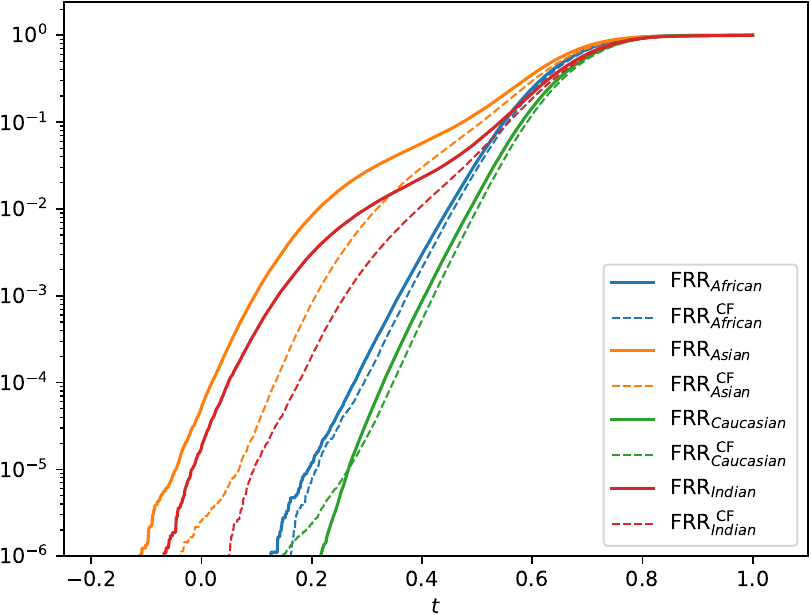}
  \captionof{figure}{Real metrics $\frr_a$ obtained with real scores, for each race. The real scores come from the pre-trained model $f$ (solid lines) or the Fairness Module (dashed).}
  \label{fig:rs_frr}
\end{minipage}
\end{figure*}

\textbf{Pre-trained model.} We take as pre-trained model $f$ a ResNet100 \citep{resnet100_forFR}, which we train during $20$ epochs on BUPT with the ArcFace \citep{arcface} loss function (see \ref{app:arcface_training} for details). As many FR models, its embedding dimension is $d=512$. To train our Fairness Module efficiently, we infer the embeddings of the whole training set and save them. Those pre-trained embeddings are the input of our Fairness Module. From those embeddings, we also compute the pre-trained centroids $\vect{\mu}_k^{(0)}$ and pseudo-scores $\overline{s}(\vect{x}_i, \vect{\mu}_k^{(0)})$.

\textbf{Fairness models.} As Caucasians are often the most represented group within FR datasets and that they benefit from better performance than other subgroups \citep{rfw, cavazos2020accuracy}, we set the reference group as $r=\text{Caucasian}$. The Fairness Module is trained with $\mathcal{L}_{\mathrm{CF}}$ for $20$ epochs on the ArcFace embeddings of BUPT, with a batch size of $4096$, a learning rate equal to $10^{-3}$, using the Adam \citep{adam} optimizer, on $2$ Tesla-V100-32GB GPUs during $40$ minutes. Note that our loss function $\mathcal{L}_\mathrm{CF}$ does not have any hyperparameter. The current state-of-the-art post-processing method for racial bias mitigation of FR models is achieved by PASS-s \citep{pass}, which we train on the ArcFace embeddings (see Appendix~\ref{app:pass_training_details} for details). As our method, PASS-s transforms the embeddings generated by a pre-trained model. However, it adopts an adversarial training paradigm, simultaneously training the model to classify identities while  minimizing the encoding of race within the new embeddings. Despite the appeal of race-independent embeddings, we think that demographic characteristics are an essential part of one’s identity, and their elimination may result in a notable performance loss.

\begin{table*}[t]
\caption{Evaluation metrics on RFW at several $\far$ levels. The $\roc$ metric is expressed as a percentage (\%). $\mathbf{Bold}$=Best, \underline{Underlined}=Second best.}
\label{tab:rfw_metrics}
\begin{center}
{\footnotesize
\begin{sc}
\setlength{\tabcolsep}{2.5pt}
\begin{tabular}{lcccccccccr}
\toprule
& \multicolumn{3}{c}{$\far=10^{-6}$}  
& \multicolumn{3}{c}{$\far=10^{-5}$} 
& \multicolumn{3}{c}{$\far=10^{-4}$}  \\
\cmidrule(lr){2-4} 
\cmidrule(lr){5-7} 
\cmidrule(lr){8-10} 
Model  & 
$\roc$ (\%) & $\bfar$ & $\bfrr$ & 
$\roc$ (\%) & $\bfar$ & $\bfrr$ & 
$\roc$ (\%) & $\bfar$ & $\bfrr$  \\
\midrule
ArcFace 
& $\mathbf{23.15}$ & $3.75$ & $1.24$ 
& $\mathbf{13.39}$ & $3.13$ & $1.31$
&  $\underline{6.19}$ & $\underline{2.68}$ & $1.41$ \\
ArcFace + PASS-s 
& $43.80$ & $\underline{3.42}$ & $\mathbf{1.14}$
& $30.60$ & $\underline{2.98}$ & $\mathbf{1.20}$
& $18.62$ & $2.77$ & $\underline{1.28}$ \\
ArcFace + CF 
& $\underline{23.30}$ & $\mathbf{2.37}$ & $\underline{1.16}$
& $\underline{13.44}$ & $\mathbf{2.29}$ & $\underline{1.21}$
&  $\mathbf{6.10}$ & $\mathbf{2.04}$ & $\mathbf{1.24}$ \\
\bottomrule
\end{tabular}
\end{sc}
}
\end{center}
\vskip -0.1in
\end{table*}

\begin{table*}[t]
\caption{Evaluation metrics on FairFace at several $\far$ levels. The $\roc$ metric is expressed as a percentage (\%). $\mathbf{Bold}$=Best, \underline{Underlined}=Second best.}
\label{tab:fairface_metrics}
\begin{center}
{\footnotesize
\begin{sc}
\setlength{\tabcolsep}{2.5pt}
\begin{tabular}{lcccccccccr}
\toprule
& \multicolumn{3}{c}{$\far=10^{-4}$}  
& \multicolumn{3}{c}{$\far=10^{-3}$} 
& \multicolumn{3}{c}{$\far=10^{-2}$}  \\
\cmidrule(lr){2-4} 
\cmidrule(lr){5-7} 
\cmidrule(lr){8-10} 
Model  & 
$\roc$ (\%) & $\bfar$ & $\bfrr$ & 
$\roc$ (\%) & $\bfar$ & $\bfrr$ & 
$\roc$ (\%) & $\bfar$ & $\bfrr$  \\
\midrule
ArcFace 
& $\mathbf{26.70}$ & $3.15$ & $\underline{1.08}$ 
& $\mathbf{18.70}$ & $1.79$ & $1.11$
&  $\mathbf{11.74}$ & $1.26$ & $\underline{1.11}$ \\
ArcFace + PASS-s 
& $33.86$ & $\underline{1.75}$ & $1.10$
& $26.23$ & $\mathbf{1.39}$ & $1.11$
& $17.67$ & $\underline{1.20}$ & $1.12$ \\
ArcFace + CF 
& $\underline{28.69}$ & $\mathbf{1.51}$ & $\mathbf{1.07}$
& $\underline{19.43}$ & $\underline{1.49}$ & $\mathbf{1.09}$
&  $\underline{11.82}$ & $\mathbf{1.06}$ & $\mathbf{1.09}$ \\
\bottomrule
\end{tabular}
\end{sc}
}
\end{center}
\vskip -0.1in
\end{table*}

\textbf{Results.}
We show the pseudo-metrics $\overline{\frr}_a(t)$ and the metrics $\frr_a(t)$ for the pre-trained model $f$ on BUPT respectively in Fig.~\ref{fig:ps_frr} and \ref{fig:rs_frr} (solid lines). The analogy between both types of metrics is clear as they behave similarly and the ranking of the best performance among races is preserved. Fig.~\ref{fig:ps_frr} also displays the pseudo-metrics $\overline{\frr}_a^{\mathrm{CF}}(t)$ computed from the pseudo-scores $\overline{s}_\theta$ of the trained Fairness Module (dashed lines). As we set the Caucasians as the reference group, the loss $\mathcal{L}_{\mathrm{CF}}$ enforces the pseudo-metrics $\overline{\frr}_a^{\mathrm{CF}}(t)$ to align with $\overline{\frr}_{\text{Caucasian}}(t)$ of $f$. This is typically the case for the groups having the worst performance (Asians and Indians) while the counterpart is a degradation of the performance of the Caucasians, leading to pseudo-metrics $\overline{\frr}_a^{\mathrm{CF}}(t)$ having less bias than for $f$. This positive impact reflects on the real metrics $\frr_a^{\mathrm{CF}}(t)$ in Fig.\ref{fig:rs_frr} (dashed lines) where they become closer together than for $f$. We postpone similar observations for the $\far$ pseudo-metrics and real metrics to the supplementary material (see~\ref{app:ps_far_rs_far}). 

The pre-trained model, PASS-s and our Fairness Module (trained with Centroid Fairness loss) are evaluated on RFW in Table~\ref{tab:rfw_metrics}. Their performance ($\roc$ defined in Section~\ref{subsec:performance_FR}) as well as their fairness properties ($\bfar$, $\bfrr$ in Eq.~\eqref{eq:bfar_bfrr}) are reported, at several $\far$ levels. Our approach succeeds in reducing the bias of ArcFace in all regimes, both in terms of $\far$ and $\frr$. PASS-s is also able to mitigate this bias, most of the time. However, note that the performance of PASS-s is significantly worse than that of the pre-trained model while our Fairness Module achieves comparable performance to ArcFace. The same evaluation metrics than for RFW are reported at several $\far$ levels in Table~\ref{tab:fairface_metrics} for the FairFace dataset. Our approach succeeds in mitigating a good part of the skintone bias, while keeping similar performance to ArcFace. 
Our method is robust to a change of pre-trained model (see \ref{app:eval_other_pt_models} for three other models) and a change of training set (see~\ref{app:eval_buptef}).

\section{Conclusion}
\label{sec:conclusion}

In this paper, we provide an analogy between real FR scores and centroid-based scores. This allows to define some pseudo-metrics which bridge the gap between FR training and fairness evaluation. The Centroid Fairness loss is presented and shown to align those new metrics across subgroups, allowing a small model to reduce the pre-trained racial bias without sacrificing performance. We emphasize that no sensitive attribute is needed during deployment/inference. 

\looseness=-1 \textbf{Reproducibility.} We plan to release the code used to conduct our experiments with the Centroid Fairness loss. The training of ArcFace/PASS-s follows their official code, as specified in Appendix~\ref{app:arcface_training} and Appendix~\ref{app:pass_training_details}. 

\looseness=-1 \textbf{Impact statement.} This paper presents a novel methodology aiming at mitigating the bias of FR systems based on deep learning, when evaluated by means of a specific fairness metric described therein. We stress that the proposed approach does not fully eliminate the bias as measured by the chosen metric, nor does it address other types of bias that may be relevant in the context of practical deployments. We also stress that the purpose of our work is not to advocate for or endorse the use of FR technologies for any application in society, nor to comment on regulatory aspects. 




\clearpage

\begin{ack}
This research was partially supported by the French National Research Agency (ANR), under grant ANR-20-CE23-0028 (LIMPID project). The authors would like to thank Aurélien Bellet for his helpful comments.
\end{ack}

\bibliography{biblio}
\bibliographystyle{icml2024/icml2024}


\appendix

\newpage
\appendix
\onecolumn

\section{Additional experiments}

\subsection{$\far$ pseudo-metrics and real metrics on BUPT}\label{app:ps_far_rs_far}

\begin{figure*}[ht]
\vskip 0.2in
\centering
\begin{minipage}{.46\textwidth}
  \centering
  \includegraphics[width=1.0\columnwidth]{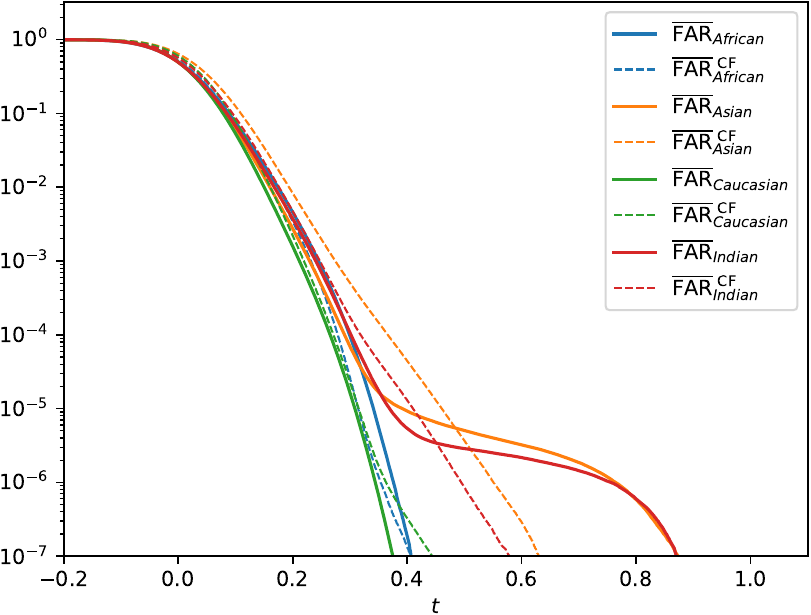}
  \captionof{figure}{Pseudo-metrics $\overline{\far}_a$ obtained with pseudo-scores, for each race. The pseudo-scores come either from the pre-trained model $f$ (solid lines), or from the Fairness Module (dashed lines).}
\label{fig:ps_far}
\end{minipage}%
\hspace{0.5cm}
\begin{minipage}{.46\textwidth}
  \centering
  \includegraphics[width=1.0\columnwidth]{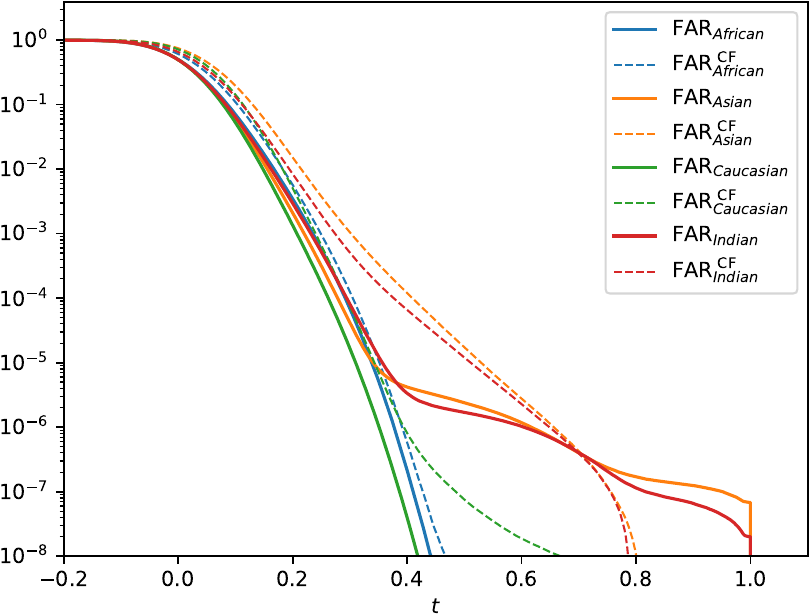}
  \captionof{figure}{Real metrics $\far_a$ obtained with real scores, for each race. The real scores come either from the pre-trained model $f$ (solid lines), or from the Fairness Module (dashed lines).}
  \label{fig:rs_far}
\end{minipage}
\vskip -0.2in
\end{figure*}

The equivalents of Fig. \ref{fig:ps_frr} and \ref{fig:rs_frr} are displayed in Fig. \ref{fig:ps_far} and \ref{fig:rs_far}. The same analogy between pseudo-metrics and real metrics can be observed (similar behavior, preserved ranking among races), for the pre-trained model $f$ and for the Fairness Module. However, our Fairness Module seems to focus its efforts on aligning the pseudo-metrics $\overline{\far}_a^{\mathrm{CF}}$ onto the reference curve $\overline{\far}_{\text{Caucasian}}$, at low $\far$ levels only. This might be due to our choice of weights $w_\far^{(i,k)}$ appearing in $\mathcal{L}_{\far}$. We observed empirically that this choice of weights tends to give too much importance at aligning $\overline{\far}_a^{\mathrm{CF}}$ at low $\far$ levels. Choosing these weights is a difficult question and we tried to provide some rational arguments about them. However, one benefit of our choices is that the gap between the dashed curves is reduced, for both pseudo-metrics and real metrics of our Fairness Module, at low $\far$ levels, where it is generally the most difficult task.   

\subsection{Evaluation on FairFace}\label{app:fairface_eval}

We test the pre-trained model ArcFace, PASS-s and our Fairness Module trained with Centroid Fairness loss on another dataset: the evaluation is performed with the FairFace dataset \cite{fairface}. It was introduced at the ECCV 2020 FairFace challenge and we use their predefined image pairs for evaluation. Contrary to RFW, FairFace is annotated with binary skintone attributes: $\mathcal{A}_{\text{FF}} = \{\text{Dark, Bright}\}$. The skintone bias is related to the racial bias as there is some correlation between race and skintone \cite{correlation_skintone_race, pass}. Note that we consider $\mathcal{A}=\{\text{African, Asian, Caucasian, Indian}\}$ for training and for evaluating on RFW, and $\mathcal{A}_{\text{FF}}$ for the FairFace evaluation.

The performance ($\roc$) of the three models, as well as their fairness properties ($\bfar$, $\bfrr$), are reported at several $\far$ levels in Table~\ref{tab:fairface_metrics_bis}. The same conclusions as for RFW (Table~\ref{tab:rfw_metrics}) can be drawn. Our approach succeeds in mitigating a good part of the racial bias, while keeping similar performance to ArcFace.

\begin{table*}[h]
\caption{Evaluation metrics on FairFace at several $\far$ levels. The $\roc$ metric is expressed as a percentage (\%). $\mathbf{Bold}$=Best, \underline{Underlined}=Second best.}
\label{tab:fairface_metrics_bis}
\vskip 0.15in
\begin{center}
\begin{small}
\begin{sc}
\begin{tabular}{lcccccccccr}
\toprule
& \multicolumn{3}{c}{$\far=10^{-4}$}  
& \multicolumn{3}{c}{$\far=10^{-3}$} 
& \multicolumn{3}{c}{$\far=10^{-2}$}  \\
\cmidrule(lr){2-4} 
\cmidrule(lr){5-7} 
\cmidrule(lr){8-10} 
Model  & 
$\roc$ (\%) & $\bfar$ & $\bfrr$ & 
$\roc$ (\%) & $\bfar$ & $\bfrr$ & 
$\roc$ (\%) & $\bfar$ & $\bfrr$  \\
\midrule
ArcFace 
& $\mathbf{26.70}$ & $3.15$ & $\underline{1.08}$ 
& $\mathbf{18.70}$ & $1.79$ & $1.11$
&  $\mathbf{11.74}$ & $1.26$ & $\underline{1.11}$ \\
ArcFace + PASS-s 
& $33.86$ & $\underline{1.75}$ & $1.10$
& $26.23$ & $\mathbf{1.39}$ & $1.11$
& $17.67$ & $\underline{1.20}$ & $1.12$ \\
ArcFace + CF 
& $\underline{28.69}$ & $\mathbf{1.51}$ & $\mathbf{1.07}$
& $\underline{19.43}$ & $\underline{1.49}$ & $\mathbf{1.09}$
&  $\underline{11.82}$ & $\mathbf{1.06}$ & $\mathbf{1.09}$ \\
\bottomrule
\end{tabular}
\end{sc}
\end{small}
\end{center}
\vskip -0.1in
\end{table*}

\subsection{Evaluations for Other Pre-Trained Models}\label{app:eval_other_pt_models}

To go beyond the pre-trained model ArcFace used in Section~\ref{sec:experiments}, a ResNet100 is trained on BUPT during $20$ epochs, with the CosFace \cite{cosface} loss. Then, we train our Fairness Module with the Centroid Fairness (CF) loss on top of the CosFace embeddings of BUPT. The hyperparameters used to train CosFace and the Fairness Module are the same as for ArcFace (Section \ref{sec:experiments}). Both models are evaluated on the RFW dataset in Table~\ref{tab:rfw_metrics_cosface}, and on FairFace (see \ref{app:fairface_eval}) in Table~\ref{tab:fairface_metrics_cosface}. 

\begin{table*}[h!]
\caption{Evaluation metrics on RFW at several $\far$ levels, for the pre-trained model CosFace. The $\roc$ metric is expressed as a percentage (\%). $\mathbf{Bold}$=Best.}
\label{tab:rfw_metrics_cosface}
\vskip 0.15in
\begin{center}
\begin{small}
\begin{sc}
\begin{tabular}{lcccccccccr}
\toprule
& \multicolumn{3}{c}{$\far=10^{-6}$}  
& \multicolumn{3}{c}{$\far=10^{-5}$} 
& \multicolumn{3}{c}{$\far=10^{-4}$}  \\
\cmidrule(lr){2-4} 
\cmidrule(lr){5-7} 
\cmidrule(lr){8-10} 
Model  & 
$\roc$ (\%) & $\bfar$ & $\bfrr$ & 
$\roc$ (\%) & $\bfar$ & $\bfrr$ & 
$\roc$ (\%) & $\bfar$ & $\bfrr$  \\
\midrule
CosFace 
& $\mathbf{21.46}$ & $4.14$ & $1.15$ 
& $12.03$ & $3.81$ & $1.20$
&  $5.30$ & $3.01$ & $1.28$ \\
CosFace + CF 
& $21.95$ & $\mathbf{3.21}$ & $\mathbf{1.10}$
& $\mathbf{11.94}$ & $\mathbf{2.82}$ & $\mathbf{1.12}$
&  $\mathbf{5.18}$ & $\mathbf{2.37}$ & $\mathbf{1.15}$ \\
\bottomrule
\end{tabular}
\end{sc}
\end{small}
\end{center}
\vskip -0.1in
\end{table*}

\begin{table*}[h]
\caption{Evaluation metrics on FairFace at several $\far$ levels, for the pre-trained model CosFace. The $\roc$ metric is expressed as a percentage (\%). $\mathbf{Bold}$=Best.}
\label{tab:fairface_metrics_cosface}
\vskip 0.15in
\begin{center}
\begin{small}
\begin{sc}
\begin{tabular}{lcccccccccr}
\toprule
& \multicolumn{3}{c}{$\far=10^{-4}$}  
& \multicolumn{3}{c}{$\far=10^{-3}$} 
& \multicolumn{3}{c}{$\far=10^{-2}$}  \\
\cmidrule(lr){2-4} 
\cmidrule(lr){5-7} 
\cmidrule(lr){8-10} 
Model  & 
$\roc$ (\%) & $\bfar$ & $\bfrr$ & 
$\roc$ (\%) & $\bfar$ & $\bfrr$ & 
$\roc$ (\%) & $\bfar$ & $\bfrr$  \\
\midrule
CosFace 
& $\mathbf{24.42}$ & $1.77$ & $1.08$ 
& $\mathbf{17.77}$ & $1.49$ & $1.08$
&  $11.13$ & $1.25$ & $1.09$ \\
CosFace + CF 
& $25.23$ & $\mathbf{1.60}$ & $\mathbf{1.07}$
& $17.78$ & $\mathbf{1.36}$ & $\mathbf{1.07}$
&  $\mathbf{10.95}$ & $\mathbf{1.14}$ & $\mathbf{1.08}$ \\
\bottomrule
\end{tabular}
\end{sc}
\end{small}
\end{center}
\vskip -0.1in
\end{table*}

Table~\ref{tab:rfw_metrics_cosface} and Table~\ref{tab:fairface_metrics_cosface} show the robustness of our method when varying the loss function used to train the pre-trained model, but keeping the same architecture (ResNet100). We now inspect the robustness when varying the architecture of the pre-trained model.

Instead of considering a ResNet100, a ResNet34 is trained on BUPT during $20$ epochs with the ArcFace loss. The hyperparameters used to train this pre-trained model ArcFace-R34 and its Fairness Module are the same as for ArcFace ResNet100 (Section \ref{sec:experiments}). Both models are evaluated on the RFW dataset in Table~\ref{tab:rfw_metrics_arcface_r34}, and on FairFace (see \ref{app:fairface_eval}) in Table~\ref{tab:fairface_metrics_arcface_r34}. 

\begin{table*}[h!]
\caption{Evaluation metrics on RFW at several $\far$ levels, for the pre-trained model ArcFace ResNet34. The $\roc$ metric is expressed as a percentage (\%). $\mathbf{Bold}$=Best.}
\label{tab:rfw_metrics_arcface_r34}
\vskip 0.15in
\begin{center}
\begin{small}
\begin{sc}
\begin{tabular}{lcccccccccr}
\toprule
& \multicolumn{3}{c}{$\far=10^{-6}$}  
& \multicolumn{3}{c}{$\far=10^{-5}$} 
& \multicolumn{3}{c}{$\far=10^{-4}$}  \\
\cmidrule(lr){2-4} 
\cmidrule(lr){5-7} 
\cmidrule(lr){8-10} 
Model  & 
$\roc$ (\%) & $\bfar$ & $\bfrr$ & 
$\roc$ (\%) & $\bfar$ & $\bfrr$ & 
$\roc$ (\%) & $\bfar$ & $\bfrr$  \\
\midrule
ArcFace-R34 
& $\mathbf{29.10}$ & $4.35$ & $1.14$ 
& $17.89$ & $4.02$ & $1.21$
&  $8.80$ & $3.14$ & $1.29$ \\
ArcFace-R34 + CF 
& $29.81$ & $\mathbf{3.58}$ & $\mathbf{1.08}$
& $\mathbf{17.80}$ & $\mathbf{3.06}$ & $\mathbf{1.11}$
&  $\mathbf{8.63}$ & $\mathbf{2.44}$ & $\mathbf{1.14}$ \\
\bottomrule
\end{tabular}
\end{sc}
\end{small}
\end{center}
\vskip -0.1in
\end{table*}

\begin{table*}[h]
\caption{Evaluation metrics on FairFace at several $\far$ levels, for the pre-trained model ArcFace ResNet34. The $\roc$ metric is expressed as a percentage (\%). $\mathbf{Bold}$=Best.}
\label{tab:fairface_metrics_arcface_r34}
\vskip 0.15in
\begin{center}
\begin{small}
\begin{sc}
\begin{tabular}{lcccccccccr}
\toprule
& \multicolumn{3}{c}{$\far=10^{-4}$}  
& \multicolumn{3}{c}{$\far=10^{-3}$} 
& \multicolumn{3}{c}{$\far=10^{-2}$}  \\
\cmidrule(lr){2-4} 
\cmidrule(lr){5-7} 
\cmidrule(lr){8-10} 
Model  & 
$\roc$ (\%) & $\bfar$ & $\bfrr$ & 
$\roc$ (\%) & $\bfar$ & $\bfrr$ & 
$\roc$ (\%) & $\bfar$ & $\bfrr$  \\
\midrule
ArcFace-R34 
& $\mathbf{29.48}$ & $1.40$ & $1.07$ 
& $\mathbf{22.21}$ & $1.28$ & $1.07$
&  $14.61$ & $1.24$ & $1.07$ \\
ArcFace-R34 + CF 
& $30.98$ & $\mathbf{1.07}$ & $\mathbf{1.05}$
& $22.32$ & $\mathbf{1.08}$ & $\mathbf{1.05}$
&  $\mathbf{14.21}$ & $\mathbf{1.03}$ & $\mathbf{1.05}$ \\
\bottomrule
\end{tabular}
\end{sc}
\end{small}
\end{center}
\vskip -0.1in
\end{table*}

Finally, a MobileFaceNet \cite{mobilefacenets} is trained on BUPT during $40$ epochs with the ArcFace loss. This architecture is much smaller than ResNet architectures. The Fairness Module is trained on BUPT during $7$ epochs using the Centroid Fairness loss. The other hyperparameters used to train this pre-trained model ArcFace-MBF and its Fairness Module are the same as for ArcFace ResNet100 (Section \ref{sec:experiments}). Both models are evaluated on the RFW dataset in Table~\ref{tab:rfw_metrics_arcface_mbf}, and on FairFace (see \ref{app:fairface_eval}) in Table~\ref{tab:fairface_metrics_arcface_mbf}. 

\begin{table*}[h!]
\caption{Evaluation metrics on RFW at several $\far$ levels, for the pre-trained model ArcFace MobileFaceNet. The $\roc$ metric is expressed as a percentage (\%). $\mathbf{Bold}$=Best.}
\label{tab:rfw_metrics_arcface_mbf}
\vskip 0.15in
\begin{center}
\begin{small}
\begin{sc}
\begin{tabular}{lcccccccccr}
\toprule
& \multicolumn{3}{c}{$\far=10^{-6}$}  
& \multicolumn{3}{c}{$\far=10^{-5}$} 
& \multicolumn{3}{c}{$\far=10^{-4}$}  \\
\cmidrule(lr){2-4} 
\cmidrule(lr){5-7} 
\cmidrule(lr){8-10} 
Model  & 
$\roc$ (\%) & $\bfar$ & $\bfrr$ & 
$\roc$ (\%) & $\bfar$ & $\bfrr$ & 
$\roc$ (\%) & $\bfar$ & $\bfrr$  \\
\midrule
ArcFace-MBF
& $\mathbf{36.65}$ & $5.32$ & $1.08$ 
& $\mathbf{24.32}$ & $4.19$ & $1.11$
&  $\mathbf{13.33}$ & $3.22$ & $1.15$ \\
ArcFace-MBF + CF 
& $37.67$ & $\mathbf{4.18}$ & $\mathbf{1.07}$
& $25.00$ & $\mathbf{3.66}$ & $\mathbf{1.09}$
&  $13.76$ & $\mathbf{2.86}$ & $\mathbf{1.10}$ \\
\bottomrule
\end{tabular}
\end{sc}
\end{small}
\end{center}
\vskip -0.1in
\end{table*}

\begin{table*}[h!]
\caption{Evaluation metrics on FairFace at several $\far$ levels, for the pre-trained model ArcFace MobileFaceNet. The $\roc$ metric is expressed as a percentage (\%). $\mathbf{Bold}$=Best.}
\label{tab:fairface_metrics_arcface_mbf}
\vskip 0.15in
\begin{center}
\begin{small}
\begin{sc}
\begin{tabular}{lcccccccccr}
\toprule
& \multicolumn{3}{c}{$\far=10^{-4}$}  
& \multicolumn{3}{c}{$\far=10^{-3}$} 
& \multicolumn{3}{c}{$\far=10^{-2}$}  \\
\cmidrule(lr){2-4} 
\cmidrule(lr){5-7} 
\cmidrule(lr){8-10} 
Model  & 
$\roc$ (\%) & $\bfar$ & $\bfrr$ & 
$\roc$ (\%) & $\bfar$ & $\bfrr$ & 
$\roc$ (\%) & $\bfar$ & $\bfrr$  \\
\midrule
ArcFace-MBF 
& $\mathbf{36.66}$ & $1.77$ & $1.06$ 
& $\mathbf{26.74}$ & $1.31$ & $\mathbf{1.05}$
&  $\mathbf{17.44}$ & $1.12$ & $1.06$ \\
ArcFace-MBF + CF 
& $48.27$ & $\mathbf{1.44}$ & $\mathbf{1.05}$
& $28.24$ & $\mathbf{1.21}$ & $\mathbf{1.05}$
&  $17.49$ & $\mathbf{1.06}$ & $\mathbf{1.05}$ \\
\bottomrule
\end{tabular}
\end{sc}
\end{small}
\end{center}
\vskip -0.1in
\end{table*}

All the results from this section confirm the robustness of our Fairness Module to a change of pre-trained model.

\subsection{Evaluations for Another Training Set}\label{app:eval_buptef}

We now investigate the robustness of our method when varying the training set used to train the pre-trained model and its Fairness Module. Note that very few open-source FR datasets have ethnicity labels. The only large-scale FR datasets that satisfy this property are BUPT-Globalface (used in Section~\ref{sec:experiments}) and BUPT-Balancedface \cite{bupt}. BUPT-Balancedface contains $1.3$M face images from $28$k celebrities and is also annotated with race attributes. Contrary to BUPT-Globalface, it is approximately race-balanced with $7$k identities for each of the four available ethnicities.

In the following, BUPT-Balancedface is employed as the training set of the pre-trained model ArcFace ResNet100 and its Fairness Module. The Fairness Module is trained during $16$ epochs with the Centroid Fairness loss, on $2$ Tesla-V100-32GB GPUs during $15$ minutes.. The framework and the remaining hyperparameters used for training are the same than for BUPT-Globalface (see Section~\ref{sec:experiments}). Both models are evaluated on the RFW dataset in Table~\ref{tab:rfw_metrics_arcface_buptef}. Those results underline the robustness of our method when varying the training set (and the distribution of its identities).

\begin{table*}[h!]
\caption{Evaluation metrics on RFW at several $\far$ levels, for the pre-trained model ArcFace ResNet100. The pre-trained model and its Fairness Module are trained on BUPT-Balancedface. The $\roc$ metric is expressed as a percentage (\%). $\mathbf{Bold}$=Best.}
\label{tab:rfw_metrics_arcface_buptef}
\vskip 0.15in
\begin{center}
\begin{small}
\begin{sc}
\begin{tabular}{lcccccccccr}
\toprule
& \multicolumn{3}{c}{$\far=10^{-6}$}  
& \multicolumn{3}{c}{$\far=10^{-5}$} 
& \multicolumn{3}{c}{$\far=10^{-4}$}  \\
\cmidrule(lr){2-4} 
\cmidrule(lr){5-7} 
\cmidrule(lr){8-10} 
Model  & 
$\roc$ (\%) & $\bfar$ & $\bfrr$ & 
$\roc$ (\%) & $\bfar$ & $\bfrr$ & 
$\roc$ (\%) & $\bfar$ & $\bfrr$  \\
\midrule
ArcFace
& $\mathbf{29.00}$ & $4.68$ & $1.29$ 
& $\mathbf{17.85}$ & $4.13$ & $1.36$
&  $\mathbf{8.78}$ & $3.39$ & $1.48$ \\
ArcFace + CF 
& $29.21$ & $\mathbf{3.72}$ & $\mathbf{1.27}$
& $17.94$ & $\mathbf{3.36}$ & $\mathbf{1.33}$
&  $8.85$ & $\mathbf{2.84}$ & $\mathbf{1.43}$ \\
\bottomrule
\end{tabular}
\end{sc}
\end{small}
\end{center}
\vskip -0.1in
\end{table*}

\section{Further Remarks}\label{app:remarks}

\subsection{Reason for Using the Same Training Set as $f$}\label{app:reason_reuse_train_set}

Note that the debiasing approach presented in this paper could work by training the fair model on another training set than the dataset used to train $f$. It would imply the inference of $f(\vect{x}_i)$ and the computation of the centroids $\vect{\mu}_k^{(0)}$ on a different training set. The whole procedure to train the fair model starts with those inputs and can be applied on any training set. However, in order to fairly compare $f$ and its post-processing fair model, the same training set is used for both models, without adding data.  

\subsection{Effect of the Pseudo-Score Transformations $T_{a \xrightarrow[]{} r}^{\far}$ and $T_{a \xrightarrow[]{} r}^{\frr}$}\label{app:effect_ps_transformation}

For simplicity, let us consider a population with two subgroups: $a$ and $r$. $r$ stands for the \textit{reference} subgroup, on which one would like to align all other subgroups. In this case, the objective is to align the curve $\overline{\far}_a(t)$ on $\overline{\far}_r(t)$, and $\overline{\frr}_a(t)$ on $\overline{\frr}_r(t)$. 

For all $(\vect{x}_i, \vect{\mu}_k^{(0)}) \in \overline{\mathcal{I}}_a$, we consider the following pseudo-score transformation for $\overline{s}_a^{(-)} :=  \overline{s}(\vect{x}_i, \vect{\mu}_k^{(0)})$:
\begin{equation*}
T_{a \xrightarrow[]{} r}^{\far}(\overline{s}_a^{(-)}) = (\overline{\far}_{r})^{-1} \circ \overline{\far}_{a} [\overline{s}_a^{(-)}].    
\end{equation*}
Assume that all pseudo-scores $\overline{s}(\vect{x}_i, \vect{\mu}_k^{(0)})$ for $(\vect{x}_i, \vect{\mu}_k^{(0)})~\in~\overline{\mathcal{I}}_a$ were transformed with this mapping. Then, $\overline{\far}_a(t)$ would be equal to:   
\begin{equation*}
 \overline{\far}_{a \xrightarrow[]{}r}(t) = \frac{1}{|\overline{\mathcal{I}}_a|} \ \sum_{(\vect{x}_i, \vect{\mu}_k^{(0)}) \in \overline{\mathcal{I}}_a} \mathbb{I}\{ T_{a \xrightarrow[]{} r}^{\far}( \overline{s}(\vect{x}_i, \vect{\mu}_k^{(0)}) ) > t \}.  
\end{equation*}
Similarly, for all $(\vect{x}_i, \vect{\mu}_k^{(0)}) \in \overline{\mathcal{G}}_a$, we set the following pseudo-score transformation for $\overline{s}_a^{(+)} :=  \overline{s}(\vect{x}_i, \vect{\mu}_k^{(0)})$:
\begin{equation*}
T_{a \xrightarrow[]{} r}^{\frr}(\overline{s}_a^{(+)}) = (\overline{\frr}_{r})^{-1} \circ \overline{\frr}_{a} [\overline{s}_a^{(+)}].    
\end{equation*}
Assume that all pseudo-scores $\overline{s}(\vect{x}_i, \vect{\mu}_k^{(0)})$ for $(\vect{x}_i, \vect{\mu}_k^{(0)})~\in~\overline{\mathcal{G}}_a$ were transformed with this mapping. The pseudo-metric $\overline{\frr}_a(t)$ would become:  
\begin{equation*}
 \overline{\frr}_{a \xrightarrow[]{} r}(t) = \frac{1}{|\overline{\mathcal{G}}_a|} \ \sum_{(\vect{x}_i, \vect{\mu}_k^{(0)}) \in \overline{\mathcal{G}}_a} \mathbb{I}\{ T_{a \xrightarrow[]{} r}^{\frr} (\overline{s}(\vect{x}_i, \vect{\mu}_k^{(0)}) ) \leq t \}.
\end{equation*}

The transformations $T_{a \xrightarrow[]{} r}^{\far}$ and $T_{a \xrightarrow[]{} r}^{\frr}$ are illustrated in Fig.~\ref{fig:schema_score_transfo}.

\begin{proposition}\label{prop:theory_alignment}
We have that:
\[
\sup_{t \in (-1,1)} \vert \overline{\far}_{a \xrightarrow[]{}r}(t)-\overline{\far}_{r}(t) \vert \leq 1/|\overline{\mathcal{I}}_a|,
\]
\[
\sup_{t \in (-1,1)} \vert \overline{\frr}_{a \xrightarrow[]{}r}(t)-\overline{\frr}_{r}(t) \vert \leq 1/|\overline{\mathcal{G}}_a|.
\]
\end{proposition}

The proof is postponed to \ref{app:proof_theory_alignment}. This result states that $T_{a \xrightarrow[]{} r}^{\far}$ (resp. $T_{a \xrightarrow[]{} r}^{\frr}$) is an impostor (resp. genuine) pseudo-score transformation which aligns the curve $\overline{\far}_a(t)$ (resp. $\overline{\frr}_a(t)$) on $\overline{\far}_r(t)$ (resp. on $\overline{\frr}_r(t)$).

\subsection{Discussion on the weights $w_\far^{(i,k)}$ and $w_\frr^{(i,k)}$}\label{app:weights_discussion}

As detailed in Section~\ref{subsec:CF_loss}, it is necessary to enforce $w_\far^{(i,k)} \propto \mathbb{I}\{y_i \neq k\} \mathbb{I}\{a_{y_i} = a_k\}$ and $w_\frr^{(i,k)} \propto \mathbb{I}\{y_i = k\} \ \mathbb{I}\{a_{y_i} = a_k\}$. 

Let $(\vect{x}_i, \vect{\mu}_k^{(0)}) \in \overline{\mathcal{G}}_{a_k}$.
From the definition of $\overline{\frr}_{a_k}$, $90\%$ of the pseudo-scores $\overline{s}(\vect{x}_i, \vect{\mu}_k^{(0)})$ satisfy $\overline{\frr}_{a_k}[\overline{s}(\vect{x}_i, \vect{\mu}_k^{(0)})] \in (10^{-1}, 10^0]$. Thus, $90\%$ of the non-zero terms of the loss $\mathcal{L}_\frr$ enforce the Fairness Module to reach $\frr$ fairness at $\frr$ levels in $(10^{-1}, 10^0]$. Similarly, $9\%$ of the pseudo-scores $\overline{s}(\vect{x}_i, \vect{\mu}_k^{(0)})$ satisfy $\overline{\frr}_{a_k}[\overline{s}(\vect{x}_i, \vect{\mu}_k^{(0)})] \in (10^{-2}, 10^{-1}]$, enforcing $\frr$ fairness at $\frr$ levels in $(10^{-2}, 10^{-1}]$. As the $\frr$ fairness should be reached at \textit{all} $\frr$ levels, the loss $\mathcal{L}_\frr$ should give the same weight for the $9\%$ than for the $90\%$ of the pseudo-scores. Note that the $9\%$ of the pseudo-scores are $10$ times less present than the $90\%$ in $\mathcal{L}_\frr$ but that the quantity $\tilde{w}_\frr^{(i,k)} := (\overline{\frr}_{a_k}[\overline{s}(\vect{x}_i, \vect{\mu}_k^{(0)})])^{-1}$ is nearly $10$ times higher for the $9\%$ than for the $90\%$. Thus, we choose to weight each $\overline{s}(\vect{x}_i, \vect{\mu}_k^{(0)})$ by this quantity $\tilde{w}_\frr^{(i,k)}$, to ensure $\frr$ fairness at all levels.

However, there is a drawback for this weighting. Consider the case of two groups $a, r \in \mathcal{A}$, $r$ having many more images than $a$ within the training set. This scenario is typical of imbalanced/realistic training sets. In addition, the reference group $r$ is set to be the more populated subgroup in our experiment (see Section~\ref{sec:experiments}). This setting implies that $|\overline{\mathcal{G}}_r| \gg |\overline{\mathcal{G}}_a|$. For the sake of simplicity, let $|\overline{\mathcal{G}}_r| = 10^6$ and $|\overline{\mathcal{G}}_a| = 10^3$. Now, consider the pair $(\vect{x}_i, \vect{\mu}_k^{(0)}) \in \overline{\mathcal{G}}_a$ which has the lowest pseudo-score, \textit{i.e.} which minimizes $\overline{\frr}_a(t)$. From the definition of $\overline{\frr}_a(t)$, this pair satisfies $\overline{\frr}_a[ \overline{s}(\vect{x}_i, \vect{\mu}_k^{(0)}) ]~=~1/| \overline{\mathcal{G}}_a |$. Similarly, let $(\vect{x}_j, \vect{\mu}_l^{(0)}) \in \overline{\mathcal{G}}_r$ be the pair satisfying $\overline{\frr}_r[ \overline{s}(\vect{x}_j, \vect{\mu}_l^{(0)}) ] = 1/| \overline{\mathcal{G}}_r |$. The previous weights for both pairs are $\tilde{w}_\frr^{(i,k)}=10^{3}$ and $\tilde{w}_\frr^{(j,l)}=10^{6}$. Note that these pairs maximize those weights among their group $a$ or $r$. One can observe a favorable weighting for all pairs in $\overline{\mathcal{G}}_r$ compared to the pairs in $\overline{\mathcal{G}}_a$. This means that the loss $\mathcal{L}_\frr$ would enforce all groups to align their performance with the reference subgroup, but some groups more than others \textit{i.e.} those which have lots of images. To counteract this effect, we impose the maximum weighting $\tilde{w}_\frr^{(i,k)}$ among a subgroup to be equal for all subgroups. All the previous considerations on the weights in $\mathcal{L}_\frr$ can be applied similarly to the weights in $\mathcal{L}_\far$ and lead us to the final weights:
\begin{align*}
    w_\far^{(i,k)} &= \frac{\mathbb{I}\{y_i \neq k\} \mathbb{I}\{a_{y_i} = a_k\}}{ |\overline{\mathcal{I}}_{a_k}| \ \overline{\far}_{a_k}[\overline{s}(\vect{x}_i, \vect{\mu}_k^{(0)})]}, \\
    w_\frr^{(i,k)} &= \frac{\mathbb{I}\{y_i = k\} \mathbb{I}\{a_{y_i} = a_k\}}{ |\overline{\mathcal{G}}_{a_k}| \ \overline{\frr}_{a_k}[\overline{s}(\vect{x}_i, \vect{\mu}_k^{(0)})]}.
\end{align*}

\subsection{Implementation Details for the training of ArcFace}\label{app:arcface_training}

For the training of ArcFace on BUPT, we follow their official implementation\footnote{https://github.com/deepinsight/insightface/tree/master/recognition/arcface\_torch}. We train a ResNet100, with $d=512$, $0.9$ as the momentum, $5\times 10^{-4}$ as the weight decay, a batch size of $256$, a learning rate equal to $10^{-1}$, for $20$ epochs, as listed within their paper \cite{arcface}.

For the MobileFaceNet version of ArcFace (see \ref{app:eval_other_pt_models}), the model is trained during $40$ epochs with weight decay equal to~$1 \times 10^{-4}$. The other parameters remain the same than for the ResNet100.

\subsection{Implementation Details of the Centroid Fairness Loss}\label{app:CF_training_details}

Using the pre-trained model $f$, we compute all its embeddings $f(\vect{x}_i)$ on the training set and store them. Those embeddings are the input of our Fairness Module and there is no need to recompute $f(\vect{x}_i)$ each time we need $g_\theta(\vect{x}_i)$. From those embeddings, we then compute the pre-trained centroids $\vect{\mu}_k^{(0)}$ and the pre-trained pseudo-scores $\overline{s}(\vect{x}_i, \vect{\mu}_k^{(0)})$. From those pseudo-scores, one gets the pre-trained pseudo-metrics $\overline{\far}_a(t)$ and $\overline{\frr}_a(t)$ for all subgroups $a \in \mathcal{A}$. Then, using the definition of their generalized inverses in \ref{app:inverse_pseudo_far_frr}, we are able to compute $\overline{\far}_r^{-1}$ and $\overline{\frr}_r^{-1}$. All those steps allow us to define the target scores for our regression task. Those steps are achieved before the Centroid Fairness loss training.

On BUPT, we set the reference group as $r=\text{Caucasian}$. The Fairness Module is trained with $\mathcal{L}_{\mathrm{CF}}$ for $20$ epochs on the ArcFace embeddings of BUPT, with a batch size of $4096$, a learning rate equal to $10^{-3}$, using the Adam \cite{adam} optimizer. Note that our loss function $\mathcal{L}_\mathrm{CF}$ does not have any hyperparameter.

In addition to those parameters, we use a certain image sampling. The probability of sampling an image/embedding $\vect{x}_i$ (so that it appears within the next batch in the loss $\mathcal{L}_{\mathrm{CF}}$) is inversely proportional to the number of images sharing the attribute $a_{y_i}$ of $\vect{x}_i$ within the training set. Then, once an image is sampled for the current batch, all the pseudo-scores $\overline{s}_\theta(\vect{x}_i, \vect{\mu}_k)$ are computed (for all $K$ centroids) and one is able to compute the loss.  

\subsection{Implementation Details for PASS-s}\label{app:pass_training_details}

For the training of PASS-s on the ArcFace embeddings of BUPT, we follow their official implementation\footnote{https://github.com/Prithviraj7/PASS} and use the hyperparameters which they provide within their paper \cite{pass}, as listed in Table~\ref{tab:pass_parameters}.

\begin{table}[t]
\caption{Hyperparameters used to train PASS-s on BUPT.}
\label{tab:pass_parameters}
\vskip 0.15in
\begin{center}
\begin{small}
\begin{sc}
\begin{tabular}{lcr}
\toprule 
Parameter  & Value
\\
\midrule
$\lambda$ & $10$ \\
$K$ & $2$ \\
$T_{fc}$ & $10000$ \\
$T_{deb}$ & $1200$ \\
$T_{atrain}$ & $30000$ \\
$T_{plat}$ & $20000$ \\
$A^*$ & $0.95$ \\
$\alpha_1$ & $10^{-2}$ \\
$\alpha_2$ & $10^{-3}$ \\
$\alpha_3$ & $10^{-4}$ \\
$T_{ep}$ & $40$ \\
$N_{ep}$ & $50$ \\
\bottomrule
\end{tabular}
\end{sc}
\end{small}
\end{center}
\vskip -0.1in
\end{table}

\section{Technical Details}

\subsection{A Note on the Generalized Inverse of the $\far$ Quantity}\label{app:inverse_far}

In \ref{subsec:performance_FR}, we introduced the $\far$ metric, defined as:
\[
 \mathrm{FAR}(t) = \frac{1}{|\mathcal{I}|} \ \sum_{(\vect{x}_i, \vect{x}_j) \in \mathcal{I}} \mathbb{I}\{ s_\theta(\vect{x}_i, \vect{x}_j) > t \},
\]
and the $\roc$ curve as $\roc \colon \alpha \in (0,1)\mapsto \frr \big[ \ \far^{-1}(\alpha) \ \big]$.

The generalized inverse of any cumulative distribution function (cdf) $\kappa(t)$ on $\mathbb{R}$ is defined as 
\begin{equation}\label{eq:def_generalized_inverse}
\kappa^{-1}(\alpha)=\inf\{t\in \mathbb{R}:\; \kappa(t)\geq \alpha  \}, \quad \text{for } \alpha \in (0,1).
\end{equation}
If $\kappa$ is a cdf, the generalized inverse $\kappa^{-1}$ is its quantile function.

Note that the quantity $\far(t)$ is not a cdf, so that its generalized inverse is not well defined. However, the opposite of $\far(t)$, the True Rejection Rate ($\trr$), is a proper cdf:
\[\trr(t)=1-\far(t)= \frac{1}{|\mathcal{I}|} \ \sum_{(\vect{x}_i, \vect{x}_j) \in \mathcal{I}} \mathbb{I}\{ s_\theta(\vect{x}_i, \vect{x}_j) \leq t \}.\]
As such, the generalized inverse $\trr^{-1}(\alpha)$ is well defined for the $\trr$ quantity with Eq.~\eqref{eq:def_generalized_inverse}.

This allows to define the generalized inverse for $\far$. Indeed, for any $\alpha \in (0,1)$ satisfying $\far(t)=\alpha$, one would get the following $\trr$
\[
\trr(t) = 1-\alpha.
\]
The threshold $t$ of interest is found using the generalized inverse of $\trr$ (as in \citet{Hsieh1996}):
\[
\far^{-1}(\alpha) \colon= \trr^{-1}(1-\alpha) = (1-\far)^{-1}(1-\alpha).
\]

\subsection{A Note on the Generalized Inverses of the $\overline{\far}_a$ and $\overline{\frr}_a$ Quantities}\label{app:inverse_pseudo_far_frr}

The generalized inverse of either $\overline{\far}_a(t)$ or $\overline{\frr}_a(t)$ is defined similarly as in \ref{app:inverse_far}.

Note that $\overline{\frr}_a(t)$ is a proper cdf, thus its generalized inverse $\overline{\frr}_a^{-1}$ is defined with Eq.~\eqref{eq:def_generalized_inverse}. For $\overline{\far}_a(t)$, the same idea than in \ref{app:inverse_far} is used. Its generalized inverse is defined using the cdf $1-\overline{\far}_a(t)$.

\subsection{Proof of Proposition \ref{prop:theory_alignment}}\label{app:proof_theory_alignment}

We start by proving the following result:
\[
\sup_{t \in (-1,1)} \vert \overline{\frr}_{a \xrightarrow[]{}r}(t)-\overline{\frr}_{r}(t) \vert \leq 1/|\overline{\mathcal{G}}_a|.
\]

By construction, the jumps of the increasing stepwise function $\overline{\frr}_{a \xrightarrow[]{}r}(t)$ occur at points included in 
\[
\{\overline{\frr}_r^{-1}(l/|\overline{\mathcal{G}}_r|): \; l=1,\; \ldots,\; |\overline{\mathcal{G}}_r|\}, 
\]
like $\overline{\frr}_{r}(t)$. Hence, we have:
\begin{equation*}
\sup_{t\in (-1,\; +1)}\left\vert \overline{\frr}_{a \xrightarrow[]{}r}(t)- \overline{\frr}_{r}(t) \right\vert=\\
\max_{1\leq l\leq |\overline{\mathcal{G}}_r| } \left\vert \overline{\frr}_{a \xrightarrow[]{}r}(\overline{\frr}_{r}^{-1}(l/|\overline{\mathcal{G}}_r|))- l/|\overline{\mathcal{G}}_r| \right\vert.
\end{equation*}

Therefore, for all $l\in \{1,\; \ldots,\; |\overline{\mathcal{G}}_r|\}$, we have
\begin{align*}
\overline{\frr}_{a \xrightarrow[]{}r}(\overline{\frr}_{r}^{-1}(l/|\overline{\mathcal{G}}_r|)) &=
 (1/|\overline{\mathcal{G}}_a|)\times \max \Bigl\{k\in \{0,\; \ldots,\; |\overline{\mathcal{G}}_a| \}:\; k/|\overline{\mathcal{G}}_a| \leq l/|\overline{\mathcal{G}}_r| \Bigr\}\\
 &= \lfloor l |\overline{\mathcal{G}}_a| / |\overline{\mathcal{G}}_r| \rfloor \ / \ |\overline{\mathcal{G}}_a|.
\end{align*}
We obtain that
$$
\left\vert \overline{\frr}_{a \xrightarrow[]{}r}(\overline{\frr}_{r}^{-1}(l/|\overline{\mathcal{G}}_r|))- l/|\overline{\mathcal{G}}_r| \right\vert < 1/|\overline{\mathcal{G}}_a|.
$$

The second result of Proposition~\ref{prop:theory_alignment} (on the $\far$ quantity) is proved with the same arguments.

\end{document}